\documentclass[lettersize,journal]{IEEEtran}
\usepackage{amsmath,amsfonts}
\usepackage{algorithmic}
\usepackage{algorithm}
\usepackage{array}
\usepackage[caption=false,font=normalsize,labelfont=sf,textfont=sf]{subfig}
\usepackage{textcomp}
\usepackage{stfloats}
\usepackage{url}
\usepackage{verbatim}
\usepackage{graphicx}
\usepackage{cite}
\usepackage{color}
\usepackage[table]{xcolor}
\usepackage{array}

\usepackage{amssymb,amsthm,bm,hyperref,multirow}
\usepackage{multirow}
\usepackage{tabularx}
\usepackage{booktabs}
\usepackage{algorithmic}
\usepackage{algorithm}
\usepackage{balance}

\RequirePackage{utfsym}

\def \eg {\emph{e.g.}}
\def \ie {\emph{i.e.}}
\def \etal {\emph{et al. }}

\begin{document}

\title{Predictive Reasoning with Augmented Anomaly Contrastive Learning for Compositional Visual Relations}

\author{Chengtai Li, Yuting He, Jianfeng Ren,~\IEEEmembership{Sensior Member,~IEEE,} Ruibin Bai,~\IEEEmembership{Sensior Member,~IEEE,} Yitian Zhao,~\IEEEmembership{Member,~IEEE,} Heng Yu,~\IEEEmembership{Senior Member,~IEEE,} Xudong Jiang,~\IEEEmembership{Fellow,~IEEE,}
\thanks{
This work was supported by Ningbo Science \& Technology Bureau under Grant 2023Z138, 2023Z237, 2024Z110 and 2024Z124.}
\thanks{C. Li, Y. He, J. Ren R. Bai and H. Yu are with the School of Computer Science, University of Nottingham Ningbo China, Ningbo 315100, China.}
\thanks{C. Li, Y. Zhao is with Institute of Biomedical Engineering, Ningbo institute of materials technology and engineering, Chinese Academy of Sciences.} 
\thanks{X. Jiang is with School of Electrical \& Electronic Engineering, Nanyang Technological University, Nanyang Link, 639798 Singapore.} 
\thanks{\textit{Corresponding author: Jianfeng Ren,  \href{mailto:jianfeng.ren@nottingham.edu.cn}{jianfeng.ren@nottingham.edu.cn}, Yitian Zhao, \href{mailto:yitian.zhao@nottingham.edu.cn}{yitian.zhao@nimte.ac.cn},}
}
}


\markboth{}{Li \MakeLowercase{\textit{et al.}}: Predictive Reasoning with Augmented Anomaly Contrastive Learning for Compositional Visual Relations}


\maketitle

\begin{abstract}
While visual reasoning for simple analogies has received significant attention, compositional visual relations (CVR) remain relatively unexplored due to their greater complexity. To solve CVR tasks, we propose Predictive Reasoning with Augmented Anomaly Contrastive Learning (PR-A$^2$CL), \ie, to identify an outlier image given three other images that follow the same compositional rules. To address the challenge of modelling abundant compositional rules, an Augmented Anomaly Contrastive Learning is designed to distil discriminative and generalizable features by maximizing similarity among normal instances while minimizing similarity between normal and anomalous outliers. More importantly, a predict-and-verify paradigm is introduced for rule-based reasoning, in which a series of Predictive Anomaly Reasoning Blocks (PARBs) iteratively leverage features from three out of the four images to predict those of the remaining one. Throughout the subsequent verification stage, the PARBs progressively pinpoint the specific discrepancies attributable to the underlying rules. Experimental results on SVRT, CVR and MC$^2$R datasets show that PR-A$^2$CL significantly outperforms state-of-the-art reasoning models.
\end{abstract}


\section{Introduction}
\IEEEPARstart{V}{isual} recognition has advanced significantly with the development of deep neural networks~\cite{he2016deep, dosovitskiy2020image,he2024dual}. Recently, the research focus has gradually shifted from visual recognition of objects and attributes~\cite{song2024centerformer,umam2024unsupervised,wu2025mg} to visual analogical reasoning of high-level concepts perceived from images and videos~\cite{malkinski2022review, pan2024joint}. Visual reasoning can be broadly categorized into the following areas~\cite{malkinski2022review}: visual question answering~\cite{jiang2023mixphm,xue2025linin}, visual commonsense reasoning~\cite{zhu2024multi, zhang2024visual,yuan2025relation}, physical reasoning~\cite{baradel2019cophy, allen2020rapid}, mathematical visual reasoning~\cite{dai2019bridging,li2025darr} and abstract visual reasoning~\cite{benny2021scale, he2023hierarchical, yang2023neural,lidsrf}. 
Among these tasks, Abstract Visual Reasoning (AVR) continues to attract significant research interest, as it remains a critical bottleneck for intelligent agents~\cite{zador2023catalyzing}, despite substantial progress in large-scale models for computer vision~\cite{kirillov2023segment} and natural language processing~\cite{kasneci2023chatgpt}. Typically, Large Language Models (LLMs) demonstrate strong generalization, achieving state-of-the-art results across diverse natural language processing~\cite{kasneci2023chatgpt} and multimodal tasks~\cite{qu2023layoutllm,xiong20253urllm}. 
But LLMs exhibit non-competitive performance on AVR tasks~\cite{camposampiero2023abstract}, revealing critical limitations in reasoning. AVR research is hence crucial to enhance modern reasoning systems and substantially improve their abstract reasoning capabilities.

\begin{figure}[!t]
   \centering
   \includegraphics[width=1\linewidth]{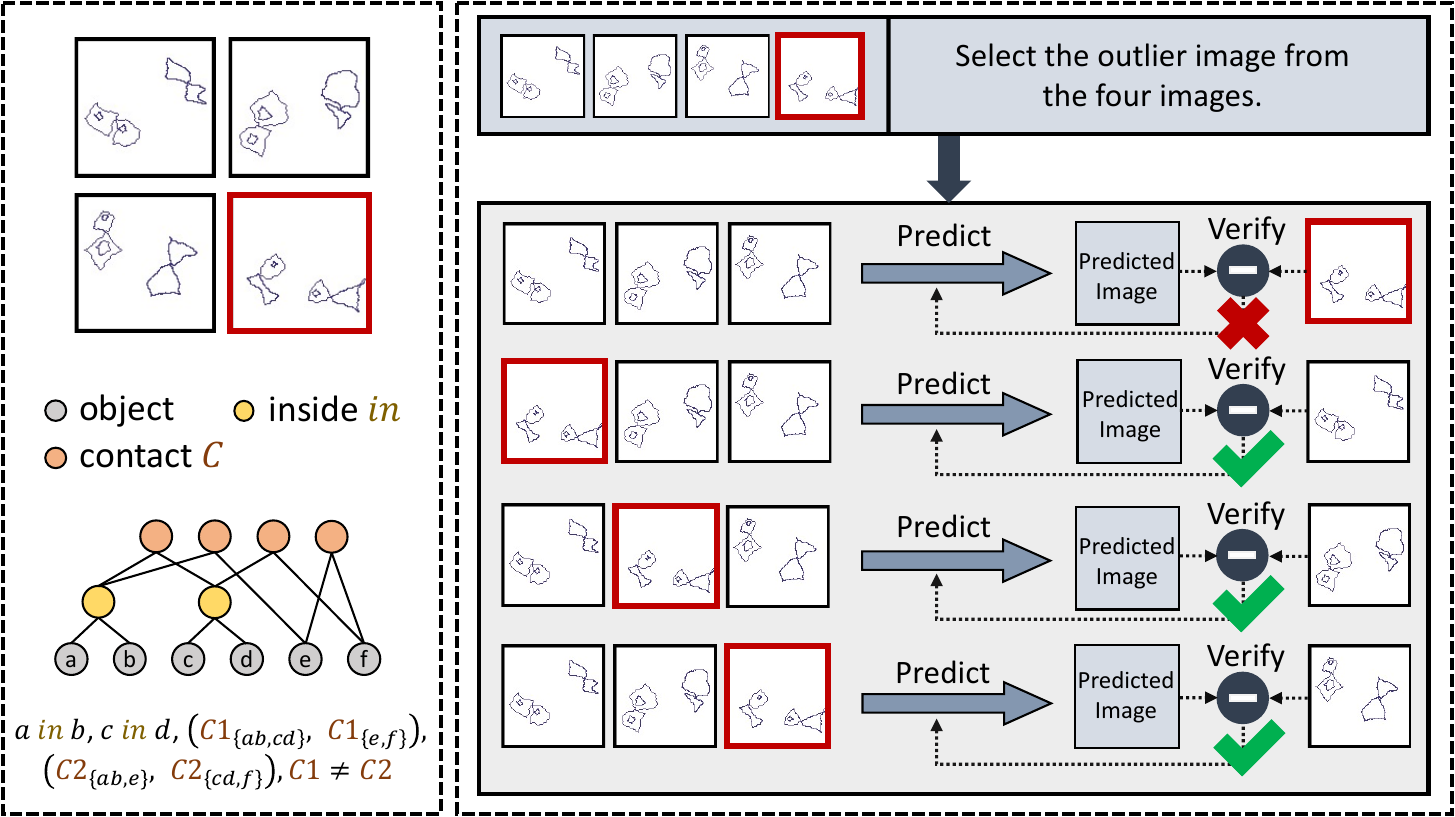}
   \caption{\textit{Left}: A sample compositional rule in the CVR dataset. 
   \textit{Right}: Selecting an outlier from four images is converted to four predict-and-verify problems. 
   }
   \label{fig:onecol}
\end{figure}

AVR research has achieved substantial progress on tasks like Raven’s Progressive Matrices (RPMs)~\cite{he2025two,wu2020scattering,he2024data,he2024hierarchical}, \eg, on RAVENs~\cite{zhang2019raven,hu2021stratified,benny2021scale}, models including SCL~\cite{wu2020scattering}, HCV-ARR~\cite{he2023hierarchical}, and PredRNet~\cite{yang2023neural} surpass human-level reasoning accuracy. 
However, these datasets often involve a limited number of attributes and rules, \eg, RAVENs~\cite{zhang2019raven,hu2021stratified,benny2021scale} include only five attributes and four simple rules. 
The recently introduced Compositional Visual Relation (CVR) dataset~\cite{zerroug2022benchmark} evaluates abstract visual reasoning by requiring models to identify an image that slightly deviates from a compositional rule shared across three other images. As shown in Fig.~\ref{fig:onecol}, compositional rules incorporate multi-layered attribute relations, in contrast to the simpler rules in RAVENs~\cite{zhang2019raven,hu2021stratified,benny2021scale}. 

In this work, we focus on solving CVR tasks~\cite{zerroug2022benchmark,li2025dbcr}, a challenging subclass of AVR. While AVR covers a broad range of relational reasoning problems, CVR specifically requires reasoning with multiple simultaneous rules and their compositional interactions. 
Specifically, CVR tasks present two major challenges:
1)~Comprehending compositional rules is significantly more difficult than understanding simple rules, as it demands integrating multiple basic attributes, modelling their interactions, and performing high-level abstractions~\cite{zerroug2022benchmark}.
2)~The potentially infinite space of compositional rules severely challenges the generalization ability of reasoning models when encountering unseen rule combinations during testing. 

To tackle these challenges, we propose \textbf{P}redictive \textbf{R}easoning with \textbf{A}ugmented \textbf{A}nomaly \textbf{C}ontrastive \textbf{L}earning (\textbf{PR}-\textbf{A$^2$CL}). It integrates a visual perception module, \textbf{A}ugmented \textbf{A}nomaly \textbf{C}ontrastive \textbf{L}earning (\textbf{A$^2$CL}) built on a pre-trained ResNet-50~\cite{zerroug2022benchmark}, to extract robust and rule-consistent features, along with a \textbf{P}redictive \textbf{A}nomaly \textbf{R}easoning \textbf{M}odule (\textbf{PARM}) that performs iterative rule inference and verification. 
Specifically, the proposed A$^2$CL improves generalization by enhancing the alignment of rule-consistent features across augmented views. It promotes high feature affinity among normal samples under diverse augmentations to capture their intrinsic characteristics, while reducing feature similarity between normal and outlier instances to increase separability. By contrasting outliers against normal samples and enforcing consistency across augmented views, A$^2$CL learns discriminative features that exhibit strong generalization  across diverse scenarios.

To handle complex compositional rules, we propose a Predict-And-Verify (PAV) paradigm. 
We first convert the task of identifying the outlier among four images into four PAV problems, as illustrated in Fig.~\ref{fig:onecol}. Each sub-problem involves predicting the features of a target image from the features of the other three, using the proposed \textbf{P}redictive \textbf{A}nomaly \textbf{R}easoning \textbf{B}lock (\textbf{PARB}). The underlying principle is that the features of an outlier cannot be reliably predicted from three normal images, whereas the features of a normal image can be accurately inferred from the other two, provided the outlier is effectively suppressed. The prediction is subsequently verified against the target, and the prediction error is back-propagated to update the prediction network. 
Upon converging, the PARB captures the visual relations between normal images and hence identifies the outlier. 
Furthermore, multiple PARBs are stacked to facilitate hierarchical reasoning over compositional rules. The initial PARB primarily captures elementary attribute-level relations, such as same \texttt{size} or same \texttt{position}, while deeper layers progressively integrate these into higher-order compositions, \eg, ``same \texttt{size} but different \texttt{shape} and spatial layout''. This architecture mirrors the multi-level structure of CVR rules and enables the model to infer abstract compositional patterns through iterative refinement.

Our contributions can be summarized as follows. 1) Our PR-A$^2$CL tackles the challenges of reasoning about compositional visual relations by extracting robust visual features and conducting abstract analogical reasoning over compositional rules. 
2) A$^2$CL extracts discriminative features by contrasting outliers against normal images while preserving feature consistency across augmented views. 
3) PARB embeds a predict-and-verify mechanism: it predicts the target image’s features using the other three images and updates itself by verifying predictions against the target. 
4) PR-A$^2$CL significantly outperforms state-of-the-art visual reasoning models on the SVRT~\cite{fleuret2011comparing}, CVR~\cite{zerroug2022benchmark} and MC$^2$R~\cite{li2024regression} datasets. 

\section{Related Work}
\subsection{Abstract Visual Reasoning}  
AVR examines the capacity for analogical reasoning on abstract visual relations across diverse scenes~\cite{malkinski2022review}. It requires applying relevant knowledge to unfamiliar scenarios and is widely used to assess human intelligence~\cite{malkinski2022review}.
In literature, AVR covers diverse tasks, \eg, Raven’s Progressive Matrices~\cite{wu2020scattering,he2024data,he2024hierarchical,he2025two}, Finding Same-different Tasks~\cite{ricci2021same,li2024regression}, Machine Number Reasoning~\cite{li2025darr} 
and Relation Game~\cite{shanahan2020explicitly}.

Among AVR tasks, RPMs have garnered significant research interest~\cite{wu2020scattering,he2024data,he2024hierarchical,he2025two}. These tasks involve extracting visual attributes to analyse row/column-wise relations and infer the missing image from candidate answers.
For example, Barrett~\etal\cite{barrett2018measuring} encoded pairwise context-context and context-multiple-choice relations by using visual features, and repeatedly applied a Relation Network to infer inter-panel relations. 
Wu~\etal\cite{wu2020scattering} developed the Scattering Compositional Learner (SCL) to extract features from object and attribute networks, and leveraged it for abstract reasoning. 
He~\etal\cite{he2023hierarchical} introduced a hierarchical ConViT to capture multi-level features, coupled with an attention-based relational reasoner for inferring underlying relations.
PredRNet employs a shallow convolutional neural network to extract visual features and directly maps the eight context images to the missing one to infer abstract rules~\cite{yang2023neural}. 
Malkinski~\etal\cite{malkinski2024one} developed a unified model for single-choice abstract visual reasoning, featuring an encoder-decoder reasoning framework and a Structure-Aware Dynamic Layer for adaptive feature organization across diverse reasoning tasks. 
Despite recent advances in AVR models~\cite{wu2020scattering,he2024data,he2024hierarchical,he2025two}, their reasoning capability remains largely confined to simple abstract rules, as many RPM benchmarks are constructed from such rules~\cite{zhang2019raven, hu2021stratified, benny2021scale}.

Compositionality has been extensively studied in  mathematics~\cite{saxton2019analysing}, logical reasoning~\cite{lake2018generalization}, and visual reasoning~\cite{thrush2022winoground}, yet remains underexplored in high-level compositional abstract reasoning. To bridge this gap, the CVR dataset~\cite{zerroug2022benchmark} was introduced to evaluate compositional understanding through flexible combinations of visual attributes. In this work, we address the challenges of compositional visual reasoning.


\subsection{Contrastive Learning}  
Contrastive learning (CL) enforces representation consistency across augmented views of instances while improving discriminability between distinct samples~\cite{son2022contrastive}. 
CL research mainly focuses on two directions. 
1)~Developing training mechanisms to enhance the generalization capability of backbones~\cite{chen2020simple, chen2021and}. 
SimCLR~\cite{chen2020simple} enforces similarity between augmented views of the same sample while maximizing their dissimilarity from other instances, and employs a learnable nonlinear transformation to enhance feature representations. 
MoCo, introduced by Chen~\etal\cite{chen2021and}, employs a queue to increase negative samples and a momentum encoder to maintain feature consistency across the queue. 
2)~Designing loss functions to better differentiate samples with varying similarity~\cite{oord2018representation, robinson2020contrastive, chuang2022robust}. 
InfoNCE loss~\cite{oord2018representation} follows the structure of binary cross-entropy (BCE) loss and incorporates a hyperparameter to modulate the penalty for hard negative samples. 
Robinson~\etal\cite{robinson2020contrastive} constructed a tunable sample distribution to prioritize hard negative samples. 
Chuang~\etal\cite{chuang2022robust} developed the Robust InfoNCE (RINCE) loss, which enforces symmetry to enhance the robustness of learned features. 
Recently, consistency learning~\cite{nie2022learning,nie2022search} improves model robustness by aligning representations or predictions across modalities, views, or data sources.
In this work, we investigate contrastive learning as a means of visual feature extraction, enabling models to acquire higher-level abstract features that exhibit greater generalizability and enhance reasoning capabilities.

\section{Proposed PR-A$^2$CL}
\subsection{Overview of Proposed PR-A$^2$CL}
As shown in Fig.~\ref{fig:onecol}, the CVR task~\cite{zerroug2022benchmark} aims to identify the outlier among four images $\{\bm{X}_i \in \mathbb{R}^{H\times W\times C}\}_{i=1}^4$. Three images are generated using the same compositional rule over a set of attributes, while the outlier is produced by a slightly different rule over the same attributes. $H$, $W$, and $C$ denote the height, width, and number of channels of images, respectively. Formally,   
\begin{equation}
\hat{\bm{y}} = \mathcal{F}_\Theta(\bm{X}_1, \bm{X}_2, \bm{X}_3, \bm{X}_4;\Theta),
\label{eq:important}
\end{equation}
where $\hat{\bm{y}} \in \mathbb{R}^4$ denotes the prediction scores, $\Theta$ denotes the model parameters and $\mathcal{F}_\Theta$ is the mapping function. 
\begin{figure*}[!t]
  \centering
   \includegraphics[width=0.96\linewidth]{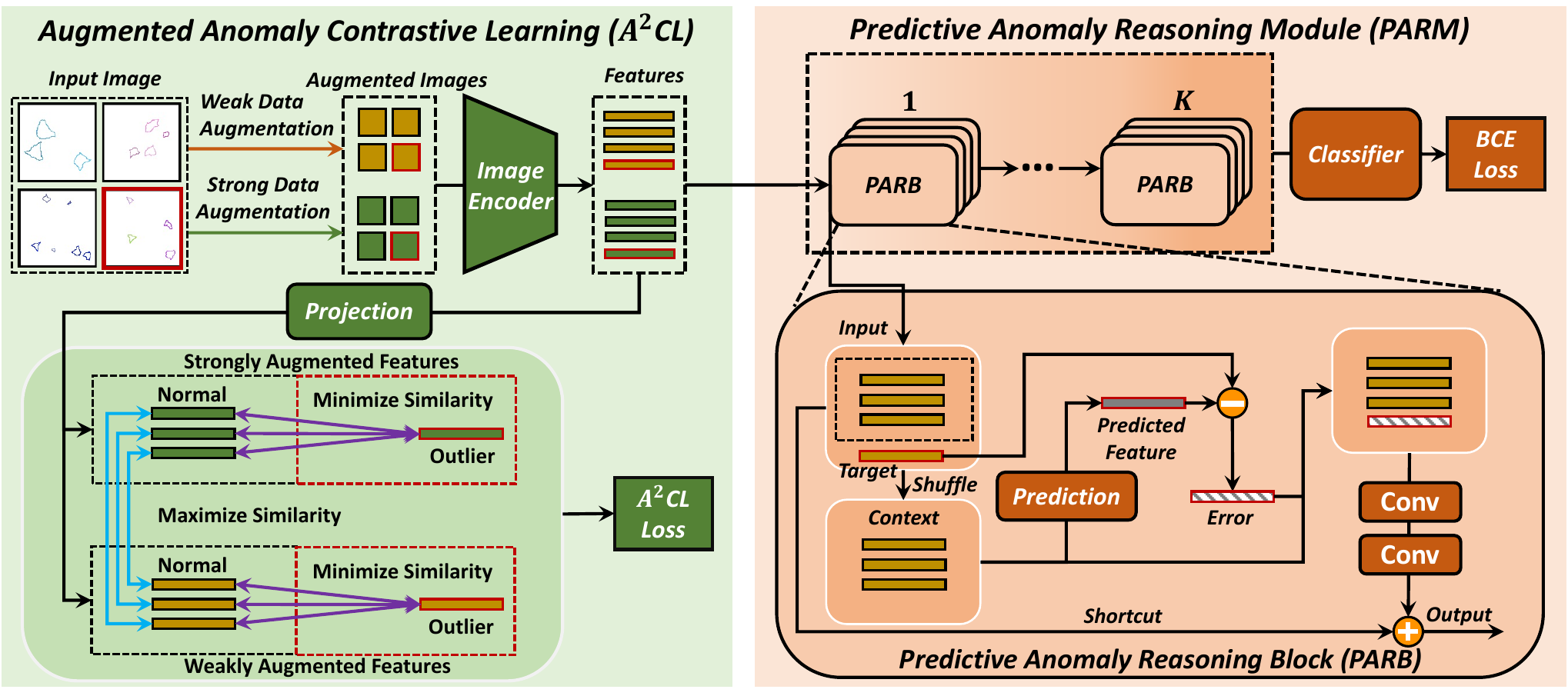}
   \caption{Proposed PR-A$^2$CL comprises two modules: 1)~A Perception Module with Augmented Anomaly Contrastive Learning (A$^2$CL), generating weak and strong augmented views, encoding them via ResNet-50, and hence improving feature discriminability and generalization. 2)~A Predictive Anomaly Reasoning Module (PARM) built with stacked PARBs. Each block employs a predict-and-verify mechanism, which predicts the target from context features and verifies it against the target to infer latent compositional rules.} 
   \label{fig:twocol}
\end{figure*}

To address the core challenges in compositional visual reasoning, we propose PR-A$^2$CL, which consists of two complementary modules, as shown in Fig.~\ref{fig:twocol}. 
Firstly, to tackle the challenge of generalizing to unseen rule compositions and appearance variations at test time, we introduce a perception module with Augmented Anomaly Contrastive Learning (A$^2$CL). It extracts robust, rule-consistent visual features by aligning representations across augmentations while separating rule-violating outliers.   
Two augmentation strategies are employed to enhance the model's generalization ability: Weak Data Augmentation (WDA) employs common image manipulations such as rotation and hue adjustment to diversify samples and expand the feature space, while Strong Data Augmentation (SDA) further improves robustness by applying masking to weakly augmented images without fundamentally altering their intrinsic attributes or semantic rules. 
Following~\cite{zerroug2022benchmark}, we adopt ResNet-50~\cite{he2016deep} as the image encoder. The A$^2$CL module is designed to minimize the similarity between outliers and normal images while maximizing agreement across augmented views, resulting in highly discriminative features with strong generalization across diverse scenarios. 

Secondly, to tackle the challenge of reasoning over compositional rules, the Predictive Anomaly Reasoning Module (PARM) employs a stacked predict-and-verify architecture.  
Specifically, the PAV paradigm transforms the outlier selection task from four images into four independent prediction-verification processes. Each process utilizes three images to predict the target image, compares the prediction with the ground truth, and updates the network parameters through prediction error minimization to implicitly learn the underlying compositional rules. 
Unlike prior models that merely predict and verify classification labels~\cite{barrett2018measuring, wu2020scattering}, this PAV paradigm is embedded within a dedicated reasoning block to enable deeper rule abstraction. 
A sequence of PARBs is applied to emulate the iterative human reasoning process, \ie, earlier layers capture elementary relations, which are progressively composed into higher-level abstract structures through subsequent PARBs.   
This hierarchical refinement allows the model to incrementally encode compositional rules, thereby facilitating the interpretation of increasingly complex rule structures.  

While PR-A$^2$CL is designed for CVR tasks~\cite{zerroug2022benchmark}, its modular architecture enables broader applicability. As shown in experiments on the SVRT dataset~\cite{fleuret2011comparing} in Sec.~\ref{sec:SVRT} and the MC$^2$R dataset~\cite{li2024regression} in Sec.~\ref{sec:MC2R}, the model effectively generalizes to multi-context compositional reasoning tasks. 


\subsection{Visual Perception with A$^2$CL} 
Human cognition can be analogized to a clustering process: even without explicit recognition of an entity, humans can associate it with others sharing similar characteristics. Integrating contrastive learning into visual reasoning frameworks enhances feature representation by emphasizing attributes relevant to underlying rules, thereby improving both discriminative power and generalization.  

\subsubsection{Weak/Strong Data Augmentation} 
Weak data augmentation comprises random rotations (90$^{\circ}$, 180$^{\circ}$, and 270$^{\circ}$), random hue adjustments, and random horizontal or vertical shifts~\cite{małkiński2024multi}. Applied with probability 
$p_w$, it enriches sample diversity by mixing weakly augmented and original data. 

Strong data augmentation operates by partitioning an image into uniformly sized blocks and applying random masking to localized regions within each block. SDA compels the model to learn from informationally sparse inputs, thereby promoting higher-level feature abstraction and enhancing robustness to image variations. To maintain consistency of the underlying semantic structure, identical augmentations are applied across all four images within a compositional reasoning sample.
 

\subsubsection{Augmented Anomaly Contrastive Learning}
The objective is to learn discriminative features that exhibit strong generalization across diverse scenarios.
Formally, let $[\bm{X}^w_1, \bm{X}^w_2, \bm{X}^w_3, \bm{O}^w]$ denote the weakly augmented images and  $[\bm{X}^s_1, \bm{X}^s_2, \bm{X}^s_3, \bm{O}^s]$ the strongly augmented counterparts, where 
 $\bm{O}^w$ and $\bm{O}^s$ represent outliers, and $\bm{X}^w_i$ and $\bm{X}^s_i$ represent augmented normal samples.   
Following~\cite{chen2020simple}, the augmented images are processed by an image encoder to extract feature representations, which are then mapped via a projection layer into an embedded space suitable for contrastive learning, 
\begin{equation}
\label{eq:weak}
[\bm{H}^w_1, \bm{H}^w_2, \bm{H}^w_3, \bm{U}^w] = \mathcal{G} \left(\mathcal{F}([\bm{X}^w_1, \bm{X}^w_2, \bm{X}^w_3, \bm{O}^w])\right), 
\end{equation}
\begin{equation}
\label{eq:strong}
[\bm{H}^s_1, \bm{H}^s_2, \bm{H}^s_3, \bm{U}^s] = \mathcal{G} \left(\mathcal{F} ([\bm{X}^s_1, \bm{X}^s_2, \bm{X}^s_3, \bm{O}^s])\right), 
\end{equation}
where $\bm{H}_i^{w}$, $\bm{H}_i^{s}$, $\bm{U}^w$ and $\bm{U}^s$ represent the features after projection, $\mathcal{F}$ represents the image encoder, and $\mathcal{G}$ represents the projection layer. 
Contrastive learning often benefits from generating diverse augmented views and maximizing the similarity among representations from these views~\cite{tian2020makes}. 
In our framework, the WDA generates diverse data views while the SDA enhances data robustness. The model employs a contrastive learning strategy that aligns features from weakly and strongly augmented normal samples, while repelling those from anomalous samples, forming a feature space where samples conforming to the same compositional rules form tight clusters to ensure semantic consistency, whereas anomalies violating rule consistency are effectively pushed away.  
Accordingly, we maximize the similarity between weakly and strongly augmented normal samples as follows, 
\begin{equation}
    \alpha_i = \mathcal{F}_{S}(\bm{H}^w_i, \bm{H}^s_i) = \frac{{\bm{H}^w_i}^T \bm{H}^s_i}{|\bm{H}^w_i| |\bm{H}^s_i|},
  \label{eq:positive-sim}
\end{equation}
where $\mathcal{F}_{S}$ is the cosine similarity. Similarity between outliers is not considered here as the outliers are diverse in nature.  
By maximizing $\alpha_i$, the feature representations derived from the weakly and strongly augmented views are encouraged to maintain high mutual similarity. 

To enhance the discriminant power, we minimize the similarity between normal and outlier samples as,
\begin{equation}
    \beta^w_i = \mathcal{F}_{S}(\bm{H}^w_i, \bm{U}^w), \quad \beta^s_i = \mathcal{F}_{S}(\bm{H}^s_i, \bm{U}^s).
  \label{eq:eg-sim2}
\end{equation}
Combining all the similarities, the A$^2$CL loss is defined as,
\begin{equation}
    \label{enq:contrastiveloss}
    \mathcal{L}_{C} = -\frac{1}{N_I}\sum^{N_I}_{i=1}{\log {\frac{\exp(\alpha_i)}{\exp(\beta^w_i)+ \exp (\beta^s_i)}}},
\end{equation}
where $N_I$ is the number of normal images. 
By minimizing $\mathcal{L}_{C}$, the learned representations promote intra-class compactness and enhance inter-class separation. This significantly improves the model’s discriminative capacity for distinguishing normal from anomalous instances, while also strengthening the generalization ability of features across diverse views.  

A$^2$CL enforces high-level structural consistency across augmented views by maximizing the similarity between weakly and strongly augmented samples while minimizing the similarity between normal and outlier instances to separate anomalies. However, it does not explicitly model compositional rule structures but instead focuses on extracting discriminative and robust features to support subsequent reasoning. 
The integration and composition of elementary relational rules into higher-order structures are achieved by PARBs via iterative predict-and-verify operations, as detailed in the next subsection. 

\subsection{Predictive Anomaly Reasoning Module}
Prior work in cognitive science indicates that human reasoning relies on iterative cycles of hypothesis generation, testing, and refinement~\cite{lake2017building,zador2023catalyzing}, \eg, Lake~\etal characterized concept learning as an iterative process of refining abstract hypotheses through feedback~\cite{lake2017building}. Zador~\etal further advocated that next-generation AI systems should integrate analogous iterative and corrective mechanisms to emulate human-like reasoning capabilities~\cite{zador2023catalyzing}. 
Inspired by cognitive science principles, our model incorporates an iterative predict-and-verify architecture that mirrors human reasoning over complex relational structures. This design is realized through stacked PARBs, in which each layer progressively refines relational rules by minimizing prediction errors. Such stepwise refinement not only enhances the accuracy of rule composition but also more closely emulates the feedback-driven mechanisms underlying human inference. 

\subsubsection{Predict-and-Verify Paradigm} 
Selecting an outlier from four images constitutes a mini-clustering task: three normal images form a coherent cluster under shared compositional rules, while the remaining one is the outlier.  However, this formulation presents two major challenges: 1) With only three normal samples and one outlier, the limited number of instances makes it difficult to reliably estimate the cluster structure; 2) Despite adhering to the same compositional rules, the three normal images may exhibit considerable visual diversity. Relying solely on feature-level similarity is therefore inadequate for capturing underlying rule consistency and ensuring semantic alignment. 
To address these challenges, we reframe the mini-clustering task within a predict-and-verify paradigm. Let $[\bm{F}_1, \bm{F}_2, \bm{F}_3, \bm{F}_4]$ denote the features of the four images. We construct four PAV instances, each taking one feature $\bm{F}_t$ as the target and the others as context to predict it, 
\begin{align}
    \label{eqn:predict}
    \hat{\bm{F}}_t = \mathcal{F}_{P}(\{\bm{F}_i\}_{i\neq t}),
\end{align}
where $\mathcal{F}_{P}$ is the prediction function. We then verify the prediction $\hat{\bm{F}}_t$ against the target as,  
\begin{align}
    \label{eqn:verify}
    \Tilde{\bm{F}}_t = \bm{F}_t- \hat{\bm{F}}_t,
\end{align}
where $\Tilde{\bm{F}}_t$ is the prediction error. 
The outlier is identified as the sample with the largest prediction error, based on the intuition that its features deviate from the shared compositional rules of normal images and thus cannot be accurately predicted from the other three. In contrast, a normal image can be reliably predicted from the remaining normal images when the outlier’s influence is suppressed. 
The prediction network acquires inherent relational patterns by predicting target features from contextual features. Through iterative optimization, the model minimizes prediction errors and continuously refines its parameters. As a result, it produces accurate predictions for normal images that conform to consistent compositional rules, while generating significant errors for anomalous samples that violate these patterns. Consequently, the prediction error strongly correlates with semantic anomaly labels. The underlying compositional rules are implicitly learned and enhanced during the iterative optimization. 
Notably, our method does not perform explicit pixel-level reconstruction from the remaining images but instead infers semantic relations and enforces compositional consistency within a structured feature space. 

In addition to the prediction formulated in~\eqref{eqn:predict}, which uses the other three images as context, we also introduce an alternative prediction scheme that relies on only two of the remaining images. This design reduces contextual noise by excluding potentially inconsistent or distracting sources, thereby promoting more focused and robust relational reasoning. Furthermore, it increases the diversity of contextual combinations during training. Both prediction strategies are integrated into our model to improve overall performance.  

\subsubsection{Predictive Anomaly Reasoning Block} 
Without loss of generality, we demonstrate one PARB that predicts the target $\bm{F}_4$ from the other three features. To ensure order invariance, $\bm{F}_1$, $\bm{F}_2$, and $\bm{F}_3$ are first shuffled. 
A reasoning network~\cite{yang2023neural} then predicts $\hat{\bm{F}}_4$ from the shuffled context: $\hat{\bm{F}}_4 = \mathcal{F}_{P}(\mathcal{F}_{\pi}(\bm{F}_1, \bm{F}_2, \bm{F}_3))$, where $\mathcal{F}_{\pi}$ denotes the permutation function. 
The network $\mathcal{F}_{P}$ is optimized by minimizing the prediction error $\Tilde{\bm{F}}_4 = \bm{F}_4 - \hat{\bm{F}}_4$, enabling it to capture underlying relations among normal images and partially learn the compositional rules when $\bm{F}_4$ is not an outlier. 
Subsequently, we concatenate $\Tilde{\bm{F}}_4$ with the context features, forming $\Tilde{\bm{Y}}_4^0 = [\bm{F}_1, \bm{F}_2, \bm{F}_3, \Tilde{\bm{F}}_4]$, which is then processed by two convolutional layers, $\mathcal{F}_{C}$, to model inter-feature interactions, $\hat{\bm{Y}}_4^0 = \mathcal{F}_{C}(\Tilde{\bm{Y}}_4^0)$. 
Additionally, a residual shortcut~\cite{he2016deep} is incorporated to leverage the original features $\bm{Y}^0 = [\bm{F}_1, \bm{F}_2, \bm{F}_3, \bm{F}_4]$, 
\begin{align}
\label{eqn:residual}
\bm{Y}_4^1 = \hat{\bm{Y}}_4^0 + \bm{Y}_4^0.      
\end{align}
The other three PARBs will obtain $\bm{Y}_1^1$, $\bm{Y}_2^1$ and  $\bm{Y}_3^1$ by choosing $\bm{F}_1$, $\bm{F}_2$ and $\bm{F}_3$ as the target, respectively.

\subsubsection{Hierarchical PARB} 
To model complex compositional rules, PARBs are stacked $K$ times hierarchically, with each PARB taking the output of the previous one as the input, 
\begin{align}
    \label{eqn:HPARB}
    \bm{Y}_i^j = \mathcal{F}_\textsc{PARB}^j(\bm{Y}_i^{j-1}), \quad 
 i=1,2,3,4,      
\end{align}
where $\mathcal{F}_\textsc{PARB}^j$ denotes the operation of the $j$-th PARB and $\bm{Y}_i^0 = \bm{Y}^0$. 
This iterative reasoning mechanism closely emulates human cognitive processes by executing cycles of prediction and verification, thereby progressively refining the inferred abstract rules~\cite{yang2023neural}.
Despite its expressive power, PARM may struggle with compositional rules involving subtle visual transformations or conflicting overlapping rules. In such cases, prediction errors may lack discriminativity between normal and anomalous images, leading to potential failures in detecting outliers, as discussed in Sec.~\ref{sec:failure_CVR}. 

Two fully connected layers $\mathcal{F}_\textsc{FC}$ then generate the outlier prediction score for the target $\bm{F}_i$, 
\begin{align}
    \label{eqn:pred_score}
    \hat{y}_i = \mathcal{F}_\textsc{FC}(\bm{Y}_i^K), \quad 
 i=1,2,3,4.       
\end{align} 
Finally, the Binary Cross Entropy (BCE) loss is adopted, 
\begin{align}
\label{eqn:bceloss}
\mathcal{L}_\textsc{BCE} = -\frac{1}{N} \sum^N_{i=1}[{\sigma(y_i) \log \sigma(\hat{y_i}) + (1-\sigma(y_i))\log(1-\sigma(\hat{y}_i))}], 
\end{align}
where $\sigma$ refers to the \texttt{sigmoid} function, $N=4$ represents the number of images in a CVR panel and $y_i$ is the one-hot encoding of ground-truth labels, \ie, $y_i=1$ if the $i$-th image is an outlier and $y_i=0$ if it is a normal image. 
The total loss of PR-A$^2$CL is defined as, 
\begin{equation}
\label{eqn:totalloss}
\mathcal{L} = \mathcal{L}_\textsc{BCE} + \lambda \mathcal{L}_{C},
\end{equation}
where $\lambda$ is the weighting factor for $\mathcal{L}_C$. 
Our model is trained in an end-to-end manner, where the network is updated by using both the contrastive loss $\mathcal{L}_{C}$ and the BCE loss $\mathcal{L}_\textsc{BCE}$. 
Algo.~\ref{algo} summarizes the training procedures of PR-A$^2$CL. 
\begin{algorithm}[!t]
\caption{Training procedures for PR-A$^2$CL}
\label{algo}
\begin{algorithmic}[1]
\renewcommand{\algorithmicrequire}{\textbf{Input:}}
\renewcommand{\algorithmicensure}{\textbf{Output:}}
    \REQUIRE Training samples
    \ENSURE PR-A$^2$CL model
    \STATE Initialize PR-A$^2$CL model
    \FOR{$i = 1$ \TO training iterations $T$}
        \STATE \textit{// Visual Perception}
        \STATE Generate weakly augmented images $\bm{H}_i^{w}$,  $\bm{U}^w$ and strongly augmented images $\bm{H}_i^{s}$, $\bm{U}^s$;
        \STATE Generate and map features into a space suitable for contrastive learning as in Eq.~\eqref{eq:weak} and Eq.~\eqref{eq:strong};
        \STATE Calculate the A$^2$CL loss $\mathcal{L}_C$ as in Eq~\eqref{eq:positive-sim},~\eqref{eq:eg-sim2} and~\eqref{enq:contrastiveloss}.  
        \STATE \textit{// Predictive Reasoning}
        \FOR{$k=1$ \TO reasoning iterations $K$}
            \FOR{$t=1$ \TO the number of images $N$}
                \STATE Predict $\bm{F}_t$ using context features as in Eq.~\eqref{eqn:predict};
                \STATE Compute the prediction error as $\Tilde{\bm{F}}_t = \bm{F}_t - \hat{\bm{F}}_t$;
                \STATE Exploit feature interaction as in Eq.~\eqref{eqn:two_con};
                \STATE Utilize a residual structure to incorporate the original feature set as in Eq.~\eqref{eqn:residual}. 
            \ENDFOR
        \ENDFOR
        \STATE Calculate BCE loss $\mathcal{L}_\textsc{BCE}$ using Eq.~\eqref{eqn:bceloss};
        \STATE Compute total loss $\mathcal{L} = \mathcal{L}_\textsc{BCE} + \lambda \mathcal{L}_{C}$;
        \STATE Update PR-A$^2$CL model via back-propagation. 
    \ENDFOR
\end{algorithmic}
\end{algorithm}

\section{Experimental Results}
\label{sec:exp}
\subsection{Experimental Settings}
Three compositional datasets are used for evaluation. 

\noindent\textbf{SVRT}~\cite{fleuret2011comparing}: 
A classic benchmark for analyzing basic compositional rules. It comprises 23 tasks, each defined by two distinct rules. Each task requires binary classification to identify the underlying rule from visual examples. These rules are constructed from seven fundamental attributes: \texttt{shape}, \texttt{position}, \texttt{size}, \texttt{rotation}, \texttt{flip}, \texttt{inside}, and \texttt{contact}. However, the original formulation relies on rigid rule matching, which simplifies the problem and limits the evaluation of reasoning capabilities. To address this, we reformulate the task as a four-choice problem following the CVR dataset~\cite{zerroug2022benchmark}, significantly increasing its difficulty and ensuring consistency with contemporary benchmarks. 

\noindent\textbf{CVR}~\cite{zerroug2022benchmark}: A synthetically generated benchmark based on compositional abstract rules, constructed from nine fundamental attributes. These include six object attributes: \texttt{shape}, \texttt{position}, \texttt{size}, \texttt{color}, \texttt{rotation}, and \texttt{flip}, and three relational attributes: \texttt{count}, \texttt{inside}, and \texttt{contact}. These attributes form nine elementary rules, which are combined to create compositional rules. The dataset comprises 103 unique rules: 20 based on a single elementary rule, 65 combining two elementary rules, and 9 incorporating more than two. 

\noindent\textbf{MC$^2$R}~\cite{li2024regression}: A challenging benchmark for abstract visual reasoning. It is designed to evaluate a model’s capacity to identify subtle distinctions within complex compositional rules, rather than relying on superficial appearance matching. Each sample consists of five images: three conforming to the same rule and two outliers exhibiting minimal deviations. Rules are constructed from nine attribute types: six object-level (\texttt{shape}, \texttt{position}, \texttt{size}, \texttt{color}, \texttt{rotation}, \texttt{flip}) and three relational (\texttt{count}, \texttt{inside}, \texttt{contact}). They are hierarchically composed using logical (\texttt{and}, \texttt{or}, \texttt{not}), relational (\texttt{=}, \texttt{>}, \texttt{<}), and arithmetic (\texttt{+}, \texttt{-}) operators in a tree-like structure. The dataset comprises 103 distinct tasks.
All three datasets, SVRT~\cite{fleuret2011comparing}, CVR~\cite{zerroug2022benchmark}, and MC$^2$R~\cite{li2024regression}, contain 10,000 training samples, 500 validation samples, and 1,000 test samples per task. Compared to CVR, MC$^2$R introduces greater attribute diversity and a multi-answer format, substantially increasing difficulty and reducing shortcut learning. 


We compare the following state-of-the-art models.  

\noindent\textbf{WReN}~\cite{barrett2018measuring} utilizes a Relation Network to model inter-feature relationships for solving RPM problems. It is adapted here for solving CVR problems. 

\noindent\textbf{SCL}~\cite{wu2020scattering} comprises an object network for object extraction, an attribute network for attribute encoding, and a relation network for inferring underlying relations.
In \textbf{SCL-ResNet-18}~\cite{wu2020scattering}, the original encoder is replaced with ResNet-18. 

\noindent\textbf{PredRNet}~\cite{yang2023neural}, a state-of-the-art model on RPMs, is adapted for CVR. It employs a four-ResBlock encoder for visual feature extraction and a Predictive Reasoning Network to model context-answer relations. 

\noindent\textbf{SCAR}~\cite{malkinski2024one} achieves dynamic adaptability via its Structure-Aware Dynamic Layer, which adjusts computations and weights based on visual task structures. It represents a state-of-the-art unified model for diverse abstract reasoning tasks.  

\noindent\textbf{R$^3$PCL~\cite{li2024regression}} performs residual regression reasoning among image triplets and incorporates pseudo-labeled contrastive learning to enhance feature discrimination and generalization.  

\noindent\textbf{DBCR~\cite{li2025dbcr}} employs a dual-branch architecture, using hierarchical regression to model intra-cluster relations and contrastive attention to capture inter-cluster differences between normal and outlier images.  

\noindent\textbf{ResNet-50}~\cite{he2016deep} and \textbf{ViT-Small}~\cite{dosovitskiy2020image} have been widely utilized for visual recognition. Following~\cite{zerroug2022benchmark}, they are selected as benchmark methods by adding two fully connected layers as the reasoner. In addition, \textbf{IN-ResNet-50}~\cite{he2016deep} and \textbf{IN-ViT-Small}~\cite{dosovitskiy2020image} pre-trained using the ImageNet dataset~\cite{deng2009imagenet}, \textbf{SSL-ResNet-50}~\cite{chen2021and} and \textbf{SSL-ViT-Small}~\cite{chen2021and} utilizing MoCo-v3~\cite{chen2021and} pre-trained with 1.03 million samples, and 
\textbf{CLIP-ResNet-50}~\cite{radford2021learning} and~\textbf{CLIP-ViT-Base}~\cite{radford2021learning} utilizing the visual encoders of CLIP~\cite{radford2021learning} are also chosen for comparison.


The default number of PARBs is $K=3$, and the weighting factor $\lambda$ for $\mathcal{L}_C$ is $0.1$. Sec.~\ref{sec:ablation} provides detailed ablations of these two parameters. The input image resolution is $128 \times 128$ pixels. 
The maximum number of training epochs is set to 100, with early stopping after 20 epochs without improvement. The Adam optimizer is utilized with an initial learning rate of 0.0001, weight decay of $1\times10^{-4}$, and a batch size of 64. 

\begin{table}[!t]
\caption{State-of-the-art comparison on SVRT~\cite{fleuret2011comparing} under the same settings as~\cite{zerroug2022benchmark}. Our method consistently and significantly outperforms all baselines across all training sample sizes.}  
\centering
\setlength{\tabcolsep}{3pt}
\begin{tabular}{lcccccccc}
    \toprule
    Model & 20 & 50 & 100 & 200 & 500 & 1000 & AUC & 10000 \\
    \midrule
    ViT-Small~\cite{dosovitskiy2020image} & 24.7 & 24.8 & 24.9 & 26.6 & 27.6 & 29.8 & 26.4 & 40.0 \\
    IN-ViT-Small~\cite{dosovitskiy2020image} & 25.6 & 25.1 & 26.2 & 26.1 & 28.7 & 33.0 & 27.5 & 92.4 \\   SCAR~\cite{malkinski2024one}&25.8&25.9&27.0&28.3&31.0&32.9&28.5&50.3\\
    WReN~\cite{barrett2018measuring} & 26.0 & 28.2 & 27.9 & 29.7 & 33.0 & 39.7 & 30.8 & 70.3\\
    SCL~\cite{wu2020scattering}  & 25.8 & 25.1 & 25.9 & 26.6 & 31.9 & 55.0 & 31.7 & 73.4 \\
    ResNet-50~\cite{he2016deep} & 29.8 & 31.5 & 32.4 & 33.4 & 41.6 & 54.0 & 37.1 & 96.0\\
    SCL-ResNet-18~\cite{wu2020scattering} & 28.2 & 31.8 & 35.8 & 41.6 & 59.5 & 68.2 & 38.2 & 95.8 \\
    PredRNet~\cite{yang2023neural} & 31.2 & 32.2 & 33.6 & 37.3 & 44.2 & 51.3 &38.3  &96.2 \\
    SSL-ViT-Small~\cite{chen2021and} & 41.7 & 43.2 & 49.4 & 55.5 & 62.7 & 67.1 & 53.3 & 85.3\\
    IN-ResNet-50~\cite{he2016deep} & 36.6 & 40.6 & 40.4 & 49.8 & 72.1 & 83.0 & 53.8 & 93.0\\
    CLIP-ResNet-50~\cite{radford2021learning} & 34.2 & 35.5 & 44.8 & 49.3 & 81.3 & 89.3 & 55.7 & 97.6\\
    CLIP-ViT-Base~\cite{radford2021learning} & 31.0 & 39.2 & 42.1 & 55.0 & 88.1 & 92.1 & 57.9 & 98.6 \\
    SSL-ResNet-50~\cite{chen2021and} & 57.5 & 62.2 & 65.0 & 70.0 & 77.1 & 83.5 & 69.2 & 97.7\\    {R$^3$PCL~\cite{li2024regression}} & {64.8} & {71.9} & {77.2} & {80.7} & {87.5} & {93.5} & {79.3} & {98.5}\\
    {DBCR~\cite{li2025dbcr}} & {\underline{65.2}} & {\underline{73.4}} & {\underline{79.5}} & {\underline{83.1}} & {\underline{89.8}} & {\underline{94.9}} & {\underline{81.0}} & {\underline{98.8}}\\
    \midrule
    {Proposed Method} & {\textbf{68.5}} & {\textbf{78.5}} & {\textbf{85.9}} & {\textbf{92.1}} & {\textbf{96.9}} & {\textbf{98.2}} & {\textbf{86.7}} & {\textbf{99.4}} \\ \hline
\end{tabular}
\label{tab:SVRT}
\end{table}
\subsection{Comparisons on SVRT Dataset}
\label{sec:SVRT}
We evaluate compared methods on the SVRT dataset, reporting results with 20, 50, 100, 200, 500, 1k, 10k training samples per task, following~\cite{zerroug2022benchmark}. 
We observe the following from Tab.~\ref{tab:SVRT}. 
1)~PR-A$^2$CL consistently outperforms all baseline methods under every experimental condition, demonstrating its efficacy in improving feature discrimination and generalization, as well as validating the strength of the proposed PARM module in abstract reasoning. 
2)~PR-A$^2$CL yields significant performance improvements over the second-best method, DBCR~\cite{li2025dbcr}, achieving margins of 3.3\%, 5.1\%, 6.4\%, 9.0\%, 7.1\%, 3.3\%, and 0.6\% across training set sizes of 20, 50, 100, 200, 500, 1k, and 10k samples per task, respectively. These consistent gains demonstrate that PR-A$^2$CL sustains a robust advantage against strong competing approaches. 
3)~PR-A$^2$CL attains accuracies of 98.2\% with 1k training samples and 99.4\% with 10k samples per task, indicating that sufficient data enables it to accurately model and reason over nearly all compositional rules present in the SVRT dataset. 
4)~Even under low-data regimes, PR-A$^2$CL maintains strong performance, achieving accuracies of 85.9\% with 100 samples and 92.1\% with 200 samples, surpassing DBCR by margins of 6.4\% and 9.0\%, respectively. These results underscore the model's robustness and generalization capacity, demonstrating its suitability for abstract reasoning under limited supervision. 

\begin{figure}[!t]
   \centering
   \includegraphics[width=1\linewidth]{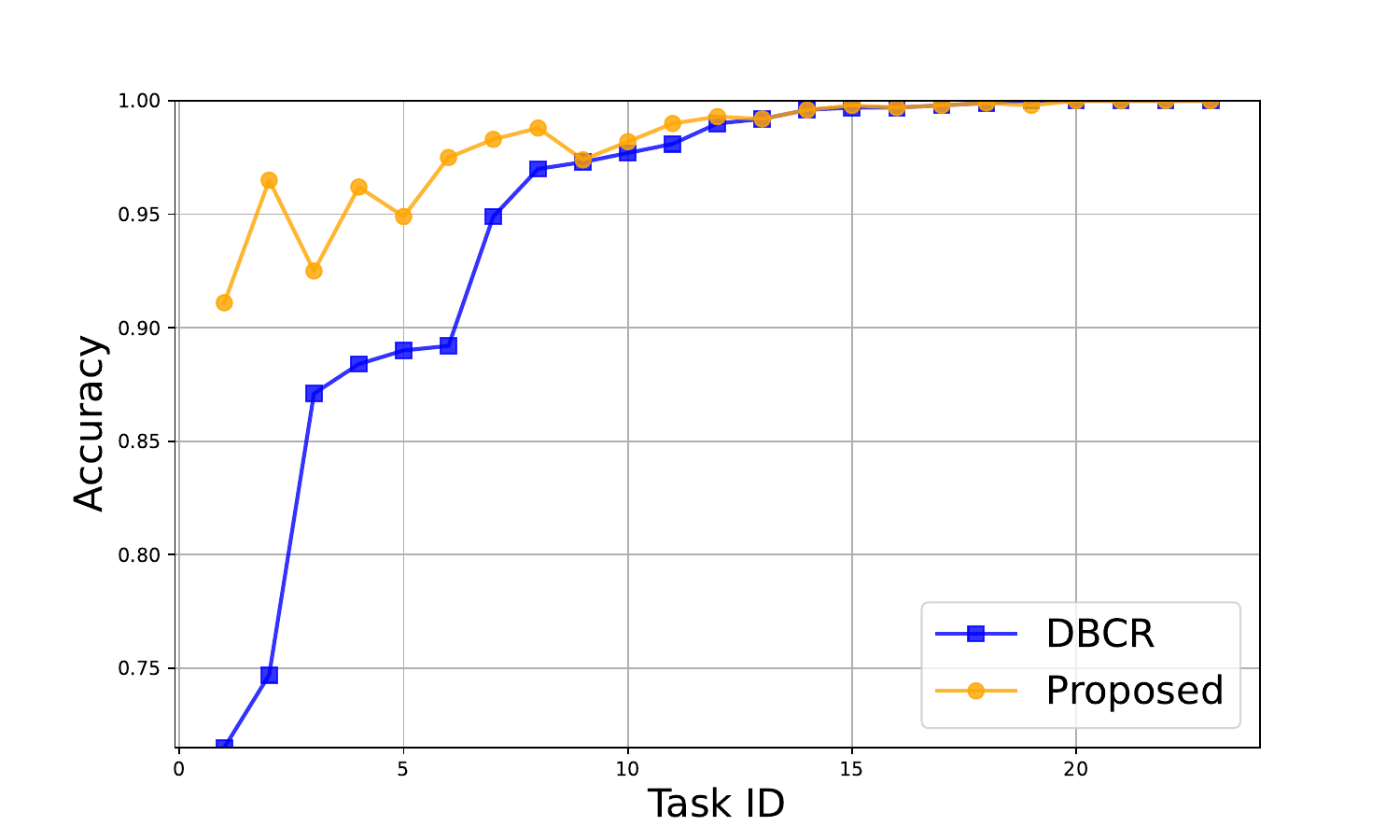}
   \caption{Comparison with DBCR~\cite{li2025dbcr} on SVRT~\cite{fleuret2011comparing} using 1k samples per task. PR-A$^2$CL excels across most tasks, whereas DBCR struggles on several.  
   }
   \label{fig:per_class_SVRT}
\end{figure}

To evaluate performance at the task level, we provide a detailed per-task accuracy comparison between PR-A$^2$CL and DBCR~\cite{li2025dbcr} under 1k training samples per task, as shown in Fig.~\ref{fig:per_class_SVRT}. Tasks are sorted in ascending order according to DBCR's accuracy.
PR-A$^2$CL achieves over 90\% accuracy on every task, highlighting its robust reasoning ability across the SVRT dataset. In contrast, DBCR exceeds 90\% on only a subset of tasks and falls below 80\% on two. We hypothesize that PR-A$^2$CL benefits from continuous self-refinement via its predict-and-verify mechanism, enabling it to handle complex rules consistently, whereas DBCR lacks such a capability.

\subsection{Comparisons on CVR Dataset}
\begin{table*}[ht]
\caption{State-of-the-art comparison on CVR~\cite{zerroug2022benchmark} under joint (\textbf{left}) and individual (\textbf{right}) training settings. PR-A$^2$CL outperforms all baselines across all sample sizes with acceptable cost. 
}
  \centering
  \setlength{\tabcolsep}{7pt}
  \begin{tabular}{@{}lccccccccrr@{}}
    \toprule
    \multirow{2}{*}{Model} & \multicolumn{8}{c}{Accuracy (\%) for Using Different Numbers of Samples per Task for Joint/Individual Training} & \multirow{2}{*}{{Para. (M)}} & \multirow{2}{*}{{GFLOPs}} \\
    \cmidrule(l){2-9}
     & 20 & 50 & 100 & 200 & 500 & 1000 & AUC & 10000 \\
    \midrule
    {SCAR~\cite{malkinski2024one}} & {26.2/27.1} & {26.2/27.4} & {26.6/27.6} & {27.1/28.0} & {27.7/28.0} & {29.4/32.3} & {27.4/28.4} & {42.7/41.4} & {0.5} & {0.7} \\
    ViT-Small~\cite{dosovitskiy2020image} & 27.3/28.6 & 27.8/30.1 & 28.0/30.9 & 28.1/31.9 & 29.9/33.8 & 31.4/35.1 & 28.7/31.7 & 58.7/42.5 & {22.6} & {11.2} \\
    WReN~\cite{barrett2018measuring} & 26.8/30.0 & 27.6/32.0 & 28.5/32.9 & 30.1/34.1 & 36.4/36.3 & 42.3/39.0 & 32.0/34.1 & 64.5/52.5 & {1.5} & {0.3}\\
    SCL~\cite{wu2020scattering}  & 25.8/26.9 & 25.8/30.0 & 28.3/30.3 & 34.1/30.0 & 43.2/31.4 & 46.2/33.4 & 33.9/30.3 & 56.9/53.2 & {0.2} & {0.1}\\
    IN-ViT-Small~\cite{dosovitskiy2020image} & 27.9/31.7 & 28.2/33.6 & 28.6/35.3 & 30.0/37.8 & 35.6/40.3 & 47.2/43.6 & 32.9/37.0 & 72.4/60.5 & {22.6} & {11.2} \\
    PredRNet~\cite{yang2023neural} & 26.9/28.9 & 30.0/32.6 & 31.5/35.3 & 36.2/39.8 & 45.9/45.3 & 54.5/49.0 & 37.5/38.5 & 92.2/68.7 & {8.4} & {5.8}\\
    ResNet-50~\cite{he2016deep} & 27.5/28.0 & 28.2/31.1 & 29.9/32.5 & 33.9/34.0 & 52.1/38.7 & 59.2/44.8 & 38.4/34.9 &93.7/84.0 & {28.0} & {10.8}\\
    SSL-ViT-Small~\cite{chen2021and} & 39.3/46.7 & 39.5/51.6 & 40.8/54.8 & 44.1/57.5 & 53.3/62.0 & 60.7/65.5 & 46.3/56.4 & 81.6/76.1 & {22.6} & {11.2}\\
    SCL-ResNet-18~\cite{wu2020scattering} & 26.4/31.4 & 28.4/37.3 & 31.6/37.8 & 40.7/39.6 & 51.4/42.7 & 64.0/48.3 & 40.4/39.5 & 78.9/72.8 & {11.5} & {4.8}\\
    IN-ResNet-50~\cite{he2016deep} & 32.0/33.9 & 35.1/36.9 & 39.0/42.8 & 43.8/48.3 & 57.7/55.3 & 69.5/62.5 & 46.2/46.6 & 85.1/82.0 & {28.0} & {10.8}\\
    CLIP-ResNet-50~\cite{radford2021learning} & 28.7/34.0 & 32.0/37.9 & 40.8/43.2 & 46.9/57.1 & 59.7/57.9 & 74.4/60.8 & 47.1/48.5 & 82.7/75.1 & {28.0} & {10.8}\\
    CLIP-ViT-Base~\cite{radford2021learning} & 31.1/33.9 & 37.4/45.7 & 43.9/49.4 & 56.0/53.7 & 68.9/54.8 & 78.8/56.1 & 52.7/48.9 & 88.9/74.6 & {22.6} & {11.2}\\
    SSL-ResNet-50~\cite{chen2021and} & 44.3/40.5 & 50.3/47.3 & 55.3/52.9 & 59.5/56.8 & 68.9/61.9 & 79.2/67.7 & 59.6/54.5 & 93.1/90.8 & {28.0} & {10.8}\\
    {R$^3$PCL~\cite{li2024regression}} & {44.8/40.9} & {52.1/49.2} & {57.6/54.1} & {63.0/60.1} & {76.4/71.1} & {88.9/86.2} & {63.8/60.3} & {95.7/93.9} & {29.6} & {12.7} \\
    {DBCR~\cite{li2025dbcr}} & {\underline{46.5}/\underline{45.3}} & {\underline{53.0}/\underline{51.8}} & {\underline{60.0}/\underline{59.1}} & {\underline{70.2}/\underline{66.9}} & {\underline{83.4}/\underline{81.2}} & {\underline{90.4}/\underline{89.5}} & {67.3/65.6} & {\underline{96.4}/\underline{94.9}} & {73.1} & {11.3}\\
    \midrule
    Proposed Method & {\textbf{47.4}/\textbf{47.2}} & {\textbf{54.3}/\textbf{53.7}} & {\textbf{62.3}/\textbf{61.0}} & {\textbf{73.2}/\textbf{69.7}} & {\textbf{84.5}/\textbf{82.1}} & {\textbf{91.8}/\textbf{91.0}} & 
{\textbf{68.9}/\textbf{67.5}} &{\textbf{97.1}}/{\textbf{95.9}} & {27.8} & {13.3}\\
    \bottomrule
  \end{tabular}
  \label{tab:SOTA}
\end{table*}

Following the experimental protocol of~\cite{zerroug2022benchmark}, we evaluate PR-A$^2$CL against state-of-the-art methods on the CVR dataset~\cite{zerroug2022benchmark} using {20, 50, 100, 200, 500, 1k, 10k} samples per task. Experiments are conducted under two configurations: a unified model trained jointly on all tasks, and individual models trained per task. Results are reported in Tab.~\ref{tab:SOTA}. 
Most baseline results are taken from~\cite{zerroug2022benchmark, li2025dbcr}. For PredRNet~\cite{yang2023neural}, we adapted the official implementation to CVR; some individually-trained models were rerun as needed. 

\subsubsection{Jointly Training One Unified Model}
As shown in Tab.~\ref{tab:SOTA}, the following observations can be made under the joint training setting. 
1)~PR-A$^2$CL consistently and significantly outperforms all baseline methods across all sample sizes, demonstrating its effectiveness in enhancing feature discrimination, generalization, and the ability of PARM to capture latent compositional rules.
2)~Compared to the second-best method, DBCR~\cite{li2025dbcr}, the performance gain of PR-A$^2$CL increases from 0.9\% to 1.3\%, 2.3\%, and 3.0\% as the number of training samples per task grows from 20 to 50, 100, and 200, respectively. This suggests that with more data, A$^2$CL better captures visual cues and improves reasoning.
3)~With 500, 1k, and 10k samples per task, the margin over DBCR decreases to 1.1\%, 1.4\%, and 0.7\%, respectively. We conjecture that abundant data (\eg, 10k samples) allows models to approximate the underlying distribution well, narrowing performance gaps. However, collecting 10k samples per task is often impractical. Thus, prior work~\cite{zerroug2022benchmark} primarily uses 1k samples for comparison, under which PR-A$^2$CL surpasses DBCR by 1.2\%.
4)~Under limited training samples, many methods perform near chance-level accuracy (25\%), while SSL-ResNet-50~\cite{zerroug2022benchmark} (MoCo-V3 pre-trained), R$^3$PCL~\cite{li2024regression}, DBCR, and our method achieve significantly better performance, highlighting the generalization benefits of A$^2$CL. 

\subsubsection{Individually Training One Model for Each Task}
As indicated in Tab.~\ref{tab:SOTA}, PR-A$^2$CL consistently surpasses all baseline methods under this setting across varying sample sizes.
Notably, it achieves a performance gain of 1.9\% over the second-best method, DBCR~\cite{li2025dbcr}, when using only 20 or 50 samples per task, a more substantial improvement than observed in the joint training configuration.
In joint training, mixing samples from multiple tasks increases the effective quantity of training data per task. For instance, joint exposure to \texttt{color} and \texttt{size} tasks can implicitly aid performance on composite tasks such as \texttt{color} \& \texttt{size}. In contrast, individual training operates with significantly fewer samples per task. Nevertheless, our approach leverages the A$^2$CL module to generate augmented data views, facilitating more efficient reasoning and yielding greater performance gains under very low sample regimes (20 or 50 samples) compared to joint training. 
Finally, we note that models trained individually on single tasks generally underperform relative to jointly trained models, primarily due to sample scarcity. Therefore, the remainder of this paper focuses on joint training results. 
As shown in the last two columns of Tab.~\ref{tab:SOTA}, we compare the time complexity of each model. PR-A$^2$CL incurs modest computational overhead but maintains high efficiency and superior accuracy across all settings. 
Moreover, PR-A$^2$CL is parameter-efficient, containing only 27.8M parameters, comparable to ResNet-50-based models, yet much fewer than DBCR~\cite{li2025dbcr}, achieving an effective accuracy-efficiency balance. 


\subsubsection{Comparisons on Each Compositional Rule}
\begin{figure}[!t]
   \centering
   \includegraphics[width=1\linewidth]{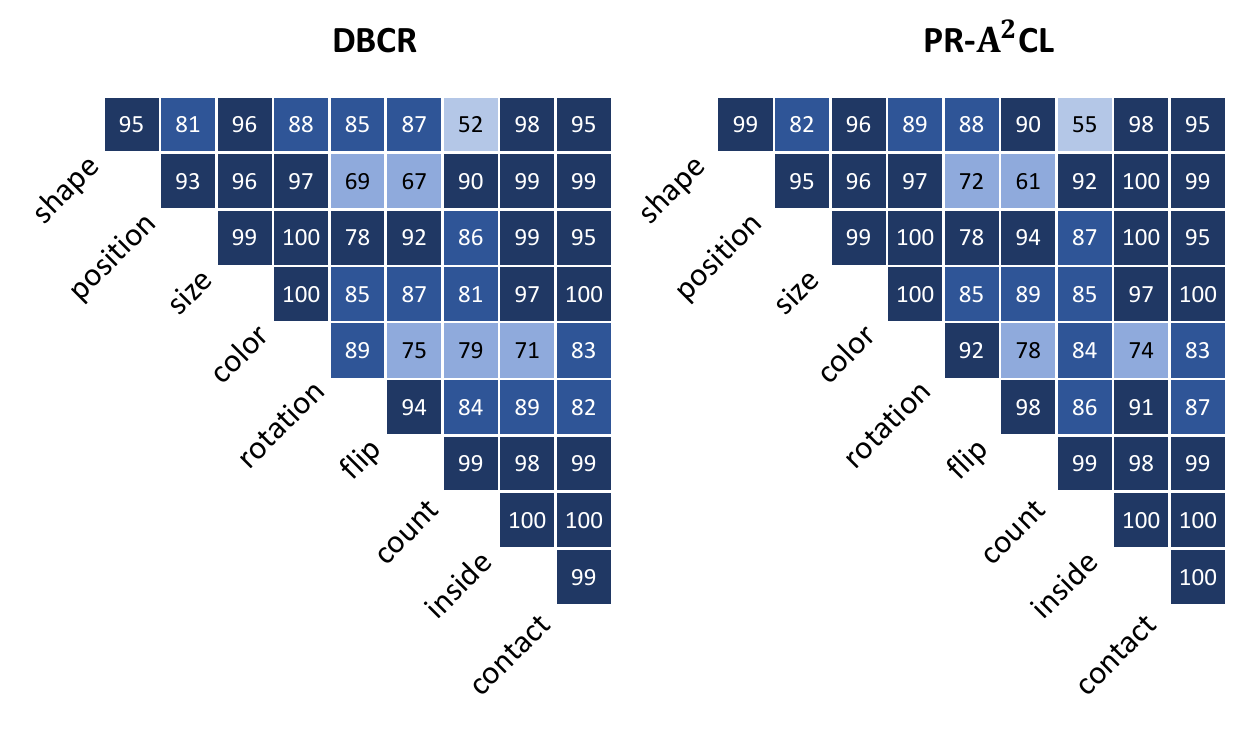}
   \caption{{Comparison of reasoning accuracy on specific compositional rules.
   PR-A$^2$CL outperforms DBCR~\cite{li2025dbcr} for almost all rules.} 
   }
   \label{fig:per_class}
\end{figure}

Fig.~\ref{fig:per_class} compares the per-rule reasoning accuracy of PR-A$^2$CL and DBCR~\cite{li2025dbcr} under 1k samples per task, averaged for rule pairs with multiple compositions. Key observations include:
1)~PR-A$^2$CL outperforms DBCR on most rule combinations, particularly in challenging compositions such as \texttt{contact+flip} and \texttt{position+flip}, indicating stronger capability in modelling multi-attribute compositions with structural dependencies.
2)~On the diagonal (elementary rules), PR-A$^2$CL achieves near-perfect accuracy across all categories, while DBCR exhibits significant degradation on abstract rules such as \texttt{shape}, \texttt{rotation}, and \texttt{flip}, suggesting limited capacity in encoding high-relational abstractions.
3)~The most substantial gains occur in semantic-sensitive tasks like \texttt{flip} and \texttt{rotation}, which require understanding of spatial structure and global context. PR-A$^2$CL’s superiority in these areas underscores its effectiveness in extracting high-level semantics and reasoning over complex visual relationships through contrastive learning and iterative predictive reasoning. 

\subsubsection{Comparisons to Human Intelligence}
Tab.~\ref{tab:human} compares the human baseline against PR-A$^2$CL across 45 abstract rules on the CVR dataset under two training regimes: the standard setting with 1k samples per task as in~\cite{zerroug2022benchmark}, and a few-shot setting with only 20 samples per task to enable fair comparison with human learners. 
While PR-A$^2$CL underperforms compared to humans under the few-shot condition, it surpasses human performance with 1k samples per task. 

Under the standard 1k samples per task setting, both humans and PR-A$^2$CL perform well on single-attribute elementary rules, but multi-rule composition remains difficult.
PR-A²CL improves accuracy from 97.6\% to 99.3\% on elementary rules and from 74.0\% to 89.1\% on compositional tasks.
It shows strong gains in spatial–semantic integrations like \texttt{Position}, \texttt{Flip}, and \texttt{Rotation} combined with other attributes. While humans struggle with limited examples (only four images per trial), PR-A$^2$CL excels via its predict-and-verify mechanism.


Under few-shot conditions, the model's accuracy on compositional rules drops sharply to 39.2\%, significantly below the human baseline of 74.0\%. The gap is largest in complex compositions such as \texttt{Color} \& \texttt{Position}, \texttt{Color} \& \texttt{Flip}, and \texttt{Count} \& \texttt{Rotation}, which require disentangling and integrating visual attributes into abstract structures. With minimal supervision, PR-A$^2$CL fails to learn robust representations, especially for subtle or rare attributes, highlighting the inherent challenge of few-shot abstract reasoning.


\begin{table}[!t]
\caption{
Human vs. PR-A$^2$CL on CVR~\cite{zerroug2022benchmark} with 20/1k samples per task, evaluated on 9 elementary and 36 compositional rules. 
}
\centering
\setlength{\tabcolsep}{5pt}
\begin{tabular}{lccc}
\toprule
& Human  & {PR-A$^2$CL, 20} & {PR-A$^2$CL, 1k} \\
\midrule
Elementary& 97.6  & {71.8} & {99.3} \\
Compositional&74.0  & {39.2} & {89.1} \\
\midrule
\texttt{Color} \& \texttt{Position}        & 57.5       & {26.2} & {98.3}                     \\
\texttt{Size} \& \texttt{Contact}               & 60.0      & {61.5} & {95.4}                     \\
\texttt{Contact} \& \texttt{Rotation}    & 30.0        & {65.1} & {94.0}                     \\
\texttt{Color} \& \texttt{Flip}  & 45.0         & {27.4}  & {88.5}                     \\
\texttt{Count} \& \texttt{Rotation}               &  45.0   &   {28.5}   & {83.5}                     \\
\bottomrule
\end{tabular}
\label{tab:human}
\end{table}

\subsection{{Comparisons on MC$^2$R Dataset}}
\label{sec:MC2R}
We summarize the key observations from Tab.~\ref{MC2R} as follows:
1)~The proposed method consistently achieves the highest performance across all experimental settings on the MC$^2$R dataset, demonstrating its strong capability in handling complex multi-context compositional reasoning tasks.
2)~Compared to the second-best method, DBCR~\cite{li2025dbcr}, our approach exhibits consistent improvements under all data regimes. Notably, in low-data scenarios with only 20, 50, and 100 training samples per task, the proposed model achieves accuracies of 22.7\%, 32.1\%, and 39.8\%, outperforming DBCR by margins of 1.3\%, 2.0\%, and 1.6\%, respectively. These improvements are especially significant given the inherent challenges of the MC$^2$R dataset, which is characterized by greater combinatorial complexity and higher rule diversity.
3)~Even with larger training sets of 1k and 10k samples, the proposed method remains highly competitive, attaining accuracies of 77.4\% and 90.4\%, respectively. These results indicate that the model not only generalizes effectively under limited supervision but also scales robustly with increased data availability. 
\begin{table}[!t]
\caption{State-of-the-art comparison on MC$^2$R~\cite{li2024regression} across varying training sample sizes. 
}
\setlength{\tabcolsep}{4pt}
\centering
    \begin{tabular}{l c c c c c c c}
    \toprule
    Model & 20 & 50 & 100 & 200 & 500 & 1,000 & 10,000 \\
    \midrule
    ViT-small~\cite{dosovitskiy2020image}     & 10.1 & 10.1 & 10.4 & 10.5 & 11.1 & 11.7 & 14.1 \\
    SCL~\cite{wu2020scattering}           & 10.1 & 10.0 & 10.1 & 10.2 & 10.5 & 12.5 & 30.4 \\
    MRNet~\cite{MRNet}       & 10.5 & 10.7 & 10.9 & 11.2 & 11.7 & 13.5 & 31.6 \\
    WReN~\cite{barrett2018measuring}     & 10.6 & 10.5 & 11.0 & 11.0 & 11.6 & 12.6 & 37.6 \\
    SSL-ViT-small~\cite{chen2021and}& 15.9 & 16.6 & 17.1 & 17.6 & 18.7 & 21.7 & 38.9 \\
    SCAR~\cite{malkinski2024one}         & 10.2 & 10.1 & 10.2 & 10.8 & 11.5 & 15.9 & 39.0 \\
    PredRNet~\cite{yang2023neural} & 11.4 & 12.6 & 13.1 & 15.6 & 23.3 & 41.2 & 59.3 \\
    ResNet-50~\cite{he2016deep}  & 10.6 & 10.5 & 10.7 & 13.1 & 17.1 & 28.1 & 75.3 \\
    SSL-ResNet-50~\cite{chen2021and}& 19.4 & 22.7 & 24.9 & 27.4 & 37.7 & 44.9 & 82.3 \\
    R$^{3}$PCL~\cite{li2024regression}  & 19.8 & 27.8 & 35.1 & 45.5 & 59.6 & 72.9 & 87.9 \\
    DBCR~\cite{li2025dbcr}           & \underline{21.4} & \underline{30.1} & \underline{38.2} & \underline{48.7} & \underline{60.8} & \underline{76.0} & \underline{89.3} \\
    \midrule
    Proposed Method & {\textbf{22.7}} & {\textbf{32.1}} & {\textbf{39.8}} & {\textbf{50.3}} & {\textbf{63.1}} & \textbf{77.4} & \textbf{90.4}\\
    \bottomrule
    \end{tabular}
\label{MC2R}
\end{table}
\subsection{Ablation Studies on CVR Dataset}
\label{sec:ablation}

\subsubsection{Ablation of Major Components} 
To assess the performance contributions of the two proposed modules, we conduct an ablation study under three data regimes: 200, 500, and 1k training samples per task. We adopt \textbf{SSL-ResNet-50}~\cite{zerroug2022benchmark} as the baseline, replacing its original perception and reasoning modules with the proposed A$^2$CL and PARM, respectively. The results are presented in Tab.~\ref{tab:major_component}. 
Integrating A$^2$CL alone leads to accuracy improvements of 6.6\%, 9.2\%, and 8.7\% across the three data scales, confirming its efficacy in learning discriminative and generalizable visual representations. Using only PARM yields gains of 4.9\%, 7.1\%, and 5.0\%, indicating that the predict-and-verify reasoning paradigm more effectively handles complex compositional rules. The combined use of both modules further elevates performance, achieving accuracies of 70.4\%, 83.9\%, and 91.5\% respectively. These results validate the individual and complementary contributions of each module to the overall framework. 



\begin{table}[!t]
\caption{Ablation study of major components on CVR~\cite{zerroug2022benchmark}.}
\centering
\setlength{\tabcolsep}{10pt}
\begin{tabular}{cc|cccc}
\toprule
A$^2$CL & PARM & 200 & 500 & 1000 \\
\midrule
\usym{2717} & \usym{2717} & 59.5 & 68.9 & 79.2 \\ 
\usym{2714} & \usym{2717} & {67.5} & {79.2} & {88.9} \\
\usym{2717} & \usym{2714} & {66.5} & {77.8}  & {85.4} \\
\usym{2714} & \usym{2714} & {\textbf{73.2}} & {\textbf{84.5}} & {\textbf{91.8}} \\
\bottomrule
\end{tabular}
\label{tab:major_component}
\end{table}
\begin{table}[!t]
\caption{{Ablation of data augmentation strategies.}} 
\centering
\setlength{\tabcolsep}{15pt}
\begin{tabular}{l|ccc}
\toprule
            & 200    & 500    & 1000    \\ 
\midrule
No CL & 66.5 & 77.8 & 85.4 \\
WDA Only & 70.8 & 82.2 & 89.1 \\
SDA Only & 67.9 & 79.4 & 87.2 \\
A$^2$CL & \textbf{73.2} & \textbf{84.5} & \textbf{91.8} \\
\bottomrule
\end{tabular}
\label{tab:data_augmentation}
\end{table}

\subsubsection{Ablation of Data Augmentation Strategies} 
To validate the efficacy of A$^2$CL, we conducted an ablation study comparing the following configurations: No Contrastive Learning (No CL), WDA Only, SDA Only, and the proposed A$^2$CL. As summarized in Tab.~\ref{tab:data_augmentation}, A$^2$CL achieves the highest reasoning accuracy across all sample sizes: 73.2\%, 84.5\%, and 91.8\% with 200, 500, and 1k training samples per class, respectively. In comparison, the absence of contrastive learning results in a significant performance decline, yielding accuracies of only 66.5\%, 77.8\%, and 85.4\%. While employing only weak or strong data augmentations surpasses the no-contrastive-learning baseline, both are outperformed by A$^2$CL. These findings indicate that contrasting strong and weak augmented views enhances the model’s ability to capture rule-consistent and perturbation-invariant semantic representations in the feature space, thereby affirming the effectiveness of our A$^2$CL.

\subsubsection{Impact of Number of PARBs} 
We investigate the impact of the number of PARBs ($K$) in the reasoning module using 1k training samples per task, as summarized in Tab.~\ref{tab:key_parameters}. The results indicate that increasing $K$ from 1 to 3 improves model performance from 91.2\% to 91.8\%, suggesting that hierarchical reasoning with multiple PARBs enhances the capture of complex compositional rules. However, increasing $K$ to 4 leads to a slight performance decline to 91.1\%, likely due to overfitting from excessive PARB stacking. This occurs when the model adapts to noise and idiosyncrasies in the training data, rather than learning generalizable reasoning patterns, causing it to focus on spurious details rather than robust abstractions. Hence, we set $K = 3$ by default. 

\begin{table}[!t]
\caption{Ablation study of key parameters $K$ and $\lambda$.} 
\centering
\color{black} 
\arrayrulecolor{black}
\setlength{\tabcolsep}{15pt}
\begin{tabular}{l|cccc}
\toprule
$K$            & 1    & 2    & 3    & 4    \\ 
\midrule
Accuracy (\%) & {91.2} & {91.4} & {\textbf{91.8}} & {91.1} \\
\midrule\midrule
$\lambda$     &0.02       & 0.05    & 0.10    & 0.20 \\ 
\midrule
Accuracy (\%) & {91.5} & {91.4} & {\textbf{91.8}} & {91.4} \\
\bottomrule
\end{tabular}
\label{tab:key_parameters}
\end{table}

\subsubsection{Impact of Weighting Factor $\lambda$} 
PR-A$^2$CL employs two loss functions: the A$^2$CL loss $\mathcal{L}C$, which enhances the discriminative power and generalization of visual features, and the BCE loss $\mathcal{L}_\textsc{BCE}$. An ablation study on the weighting factor $\lambda$ is presented in Tab.~\ref{tab:key_parameters}. 
Given that $\mathcal{L}_\textsc{BCE}$ provides direct supervisory signals from the labels and $\mathcal{L}_C$ serves as an auxiliary loss for feature enhancement, we evaluate $\lambda$ within the range of $(0.02, 0.20)$. As shown in Tab.~\ref{tab:key_parameters}, model performance remains stable across this interval, with the highest accuracy of 91.8\% achieved at $\lambda = 0.1$. This value is consequently adopted as the default. 

\subsubsection{{t-SNE Visualization of PARBs}} 
\begin{figure*}[!t]
    \centering
    \subfloat{\includegraphics[width=0.33\linewidth]{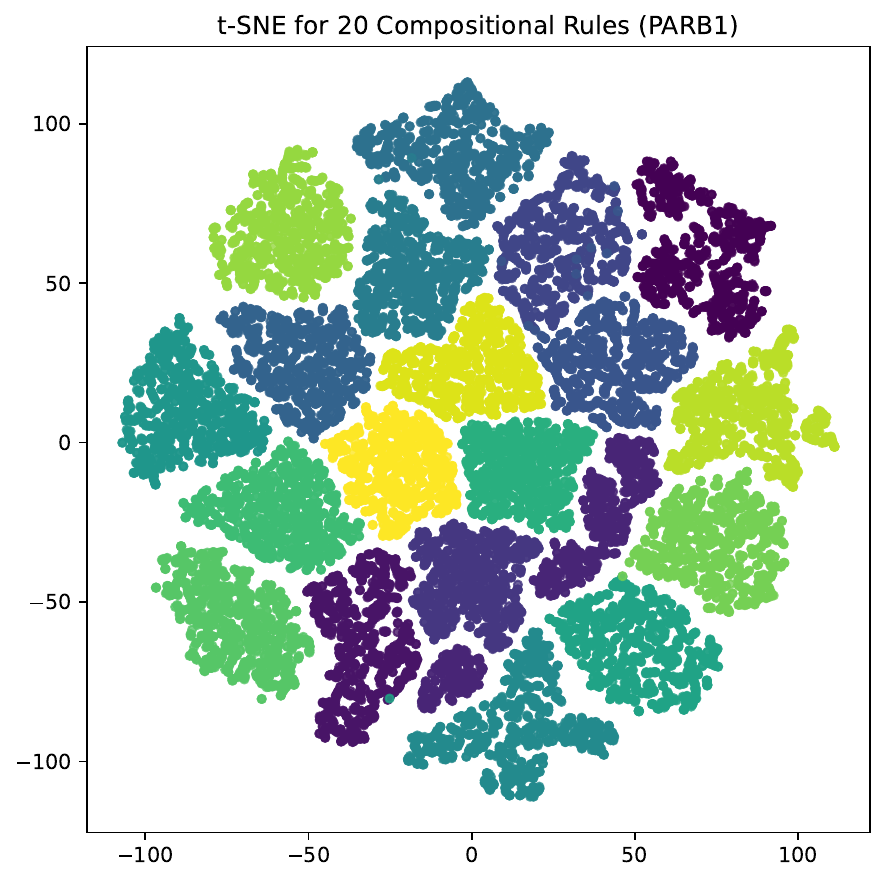}\label{fig:1a}}
    \subfloat{\includegraphics[width=0.33\linewidth]{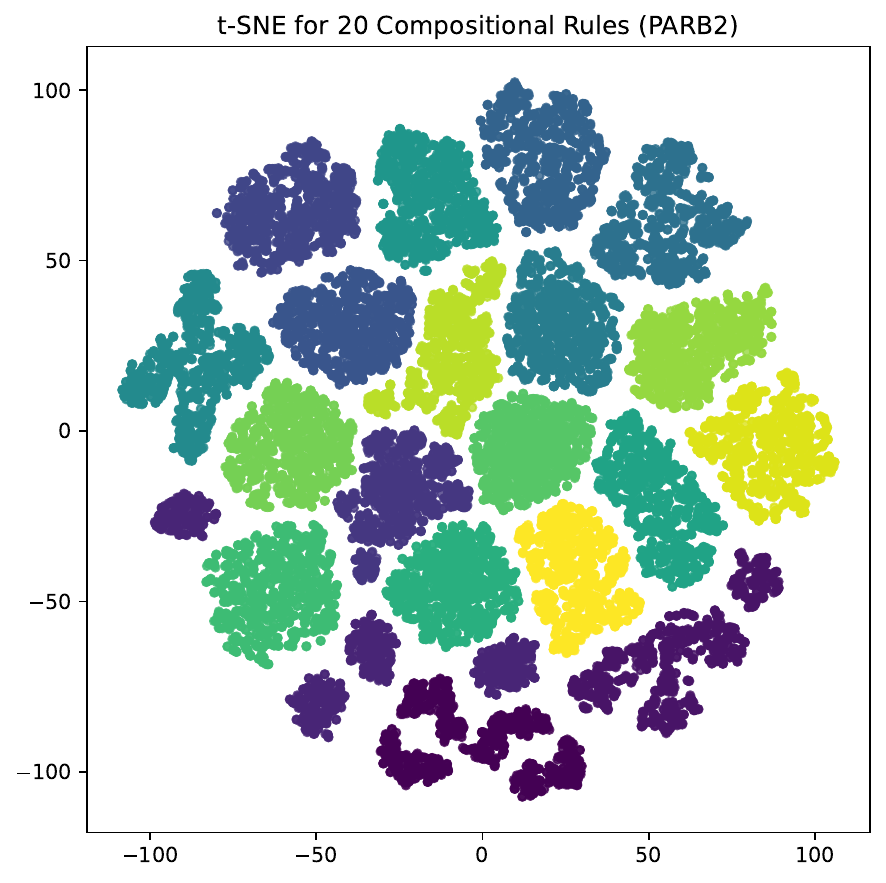}\label{fig:1b}}
    \subfloat{\includegraphics[width=0.33\linewidth]{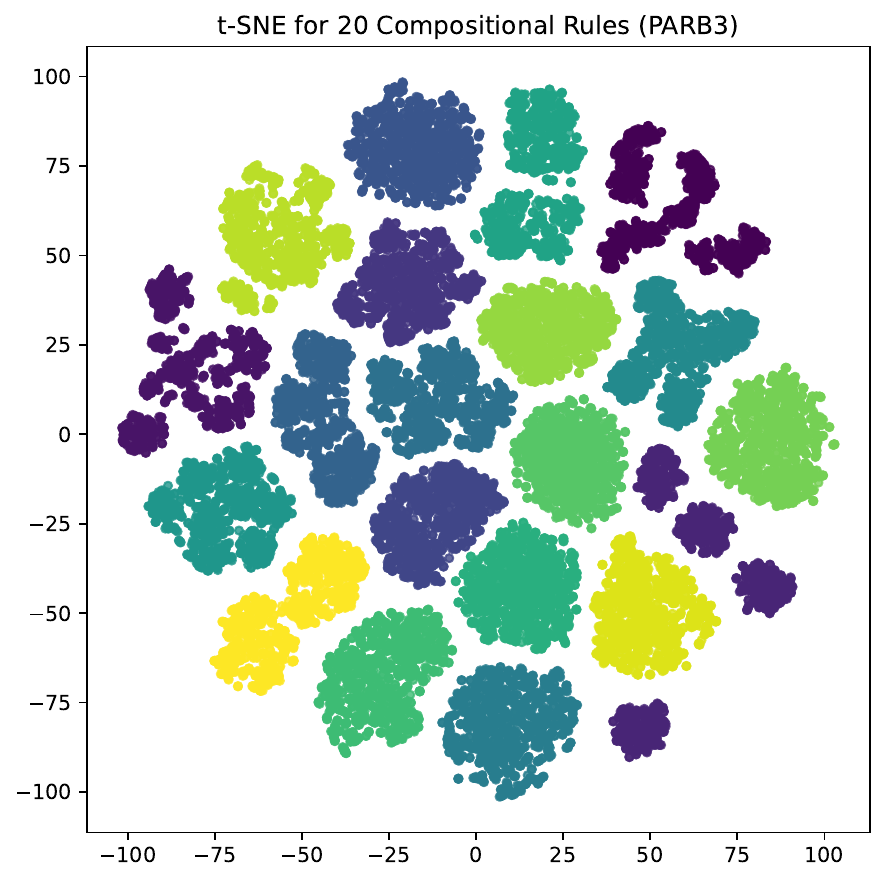}\label{fig:1c}}
    \caption{{t-SNE visualization of feature representations across successive PARBs, showing increased cluster compactness with deeper PARBs.}}
    \label{fig:t-SNE}
\end{figure*}

{To depict the evolution of feature organization throughout the iterative refinement process, we include t-SNE plots for the outputs of PARB1, PARB2 and PARB3. To ensure visual clarity and mitigate overlap, 20 compositional rules was randomly selected from the total of 103. 
As illustrated in Fig.~\ref{fig:t-SNE}, the feature representations form increasingly compact and distinct clusters across successive PARBs. This progressive separation indicates that the model refines the semantic organization within the feature space, thereby inducing a geometric configuration that more faithfully represents the underlying compositional rules.}

\subsection{Failure Analysis on CVR Dataset}
\label{sec:failure_CVR}
\begin{figure}[!t]
   \centering
   \includegraphics[width=1\linewidth]{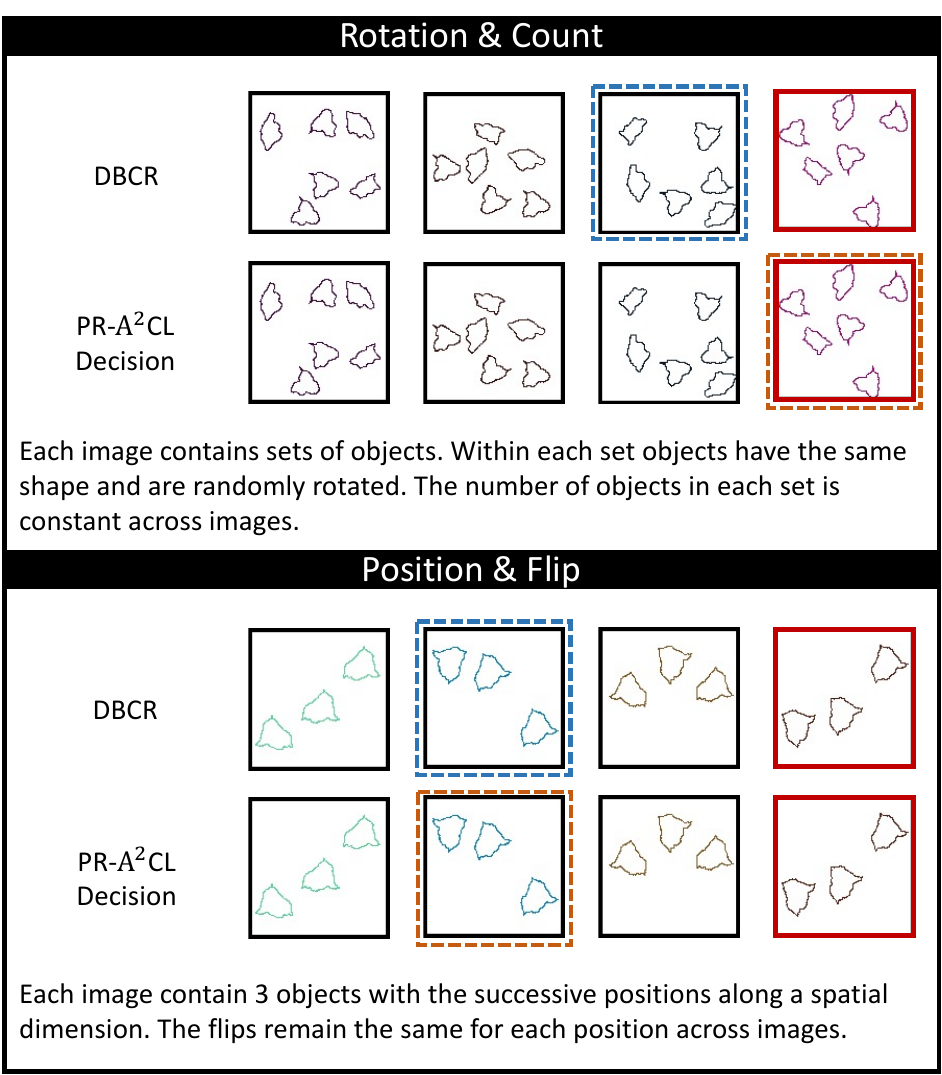}
   \caption{Two failure cases on the CVR dataset: 1)~\texttt{rotation} \& \texttt{count}: DBCR~\cite{li2025dbcr} fails to capture the specific compositional rule. In contrast, the proposed PR-A$^2$CL successfully identifies the outlier. 2)~\texttt{position} \& \texttt{flip}: Both DBCR~\cite{li2025dbcr} and PR-A$^2$CL fail to infer the underlying rule, primarily due to interference from random noise introduced by the \texttt{rotation} operation.} 
   \label{fig:fail}
\end{figure}

Fig.~\ref{fig:fail} shows two failure cases. 
On the \texttt{rotation} \& \texttt{count} task, DBCR~\cite{li2025dbcr} fails to learn the compositional rule, whereas PR-A²CL succeeds. This task involves grouping objects into subsets and identifying both the category and count of randomly rotated objects, demanding high-level semantic reasoning. While DBCR struggles with such complex compositionality, PR-A$^2$CL shows stronger reasoning over abstract visual concepts. 
The \texttt{position} \& \texttt{flip} task proves particularly challenging, as both DBCR and PR-A$^2$CL fail. This is primarily due to misinterpreting the \texttt{flip} rule, where the \texttt{rotation} acts as strong distracting noise rather than a component of the compositional rule. The rotational variability severely confuses both models, hindering accurate inference of flip-based reasoning. 
Notably, humans also struggle with these tasks, underscoring their inherent difficulty. 

Our further investigation identifies two primary failure modes:
1)~Feature attention bias. In complex task settings (\eg, \texttt{position} \& \texttt{flip}), the model exhibits a tendency to prioritize salient visual features such as object \texttt{shape} or \texttt{count}, while overlooking more subtle yet critical attributes like \texttt{flip} orientation or spatial relationships. This imbalanced attention results in systematic reasoning errors.
2)~Rule entanglement and overfitting. The model fails to decouple correlated rules (\eg, \texttt{position} \& \texttt{flip}), overfitting to superficial training patterns rather than learning abstract, generalizable rules, especially under non-additive interactions.


\section{Conclusion and Future Work}
While reasoning over simple abstract rules has been extensively explored, compositional visual reasoning remains an understudied yet critical area. This paper introduces Predictive Reasoning-Augmented Anomaly Contrastive Learning (PR-A$^2$CL), a novel framework for compositional visual reasoning. Our method includes a visual perception module enhanced with Augmented Anomaly Contrastive Learning, which extracts discriminative features by leveraging multi-level data augmentations while preserving feature-space similarity among normal samples and separating outliers. Furthermore, the predictive anomaly reasoning module employs a predict-and-verify mechanism to infer each image from the remaining three. To better understand compositional rules, we design the hierarchical PARBs to emulate human's reasoning process. Extensive experiments on SVRT, CVR, and MC$^2$R datasets demonstrate that PR-A$^2$CL consistently outperforms existing state-of-the-art models across diverse settings. 

For future work, PR-A$^2$CL can be extended by incorporating rule disentanglement or uncertainty modeling to improve performance on complex tasks. Structural feature disentanglement may yield more interpretable and independent visual attribute representations, while uncertainty-aware methods such as confidence calibration or counterfactual augmentation could enhance robustness in ambiguous or conflicting rule scenarios.

\bibliographystyle{IEEEtran}
\bibliography{main}

@STRING{PR       = "Pattern Recognit."}

@inproceedings{qu2023layoutllm,
  title={{LayoutLLM-T2I}: Eliciting layout guidance from {LLM} for text-to-image generation},
  author={Qu, Leigang and Wu, Shengqiong and Fei, Hao and Nie, Liqiang and Chua, Tat-Seng},
  booktitle={ACM MM},
  pages={643--654},
  year={2023}
}

@inproceedings{he2016deep,
  title={Deep residual learning for image recognition},
  author={He, Kaiming and Zhang, Xiangyu and Ren, Shaoqing and Sun, Jian},
  booktitle={CVPR},
  pages={770--778},
  year={2016}
}

@inproceedings{dosovitskiy2020image,
  title={An Image is Worth 16x16 Words: Transformers for Image Recognition at Scale},
  author={Dosovitskiy, Alexey and Beyer, Lucas and Kolesnikov, Alexander and Weissenborn, Dirk and Zhai, Xiaohua and Unterthiner, Thomas and Dehghani, Mostafa and Minderer, Matthias and Heigold, Georg and Gelly, Sylvain and others},
  booktitle={ICLR},
  year={2020}
}

@article{zerroug2022benchmark,
  title={A benchmark for compositional visual reasoning},
  author={Zerroug, Aimen and Vaishnav, Mohit and Colin, Julien and Musslick, Sebastian and Serre, Thomas},
  journal={NeurIPS},
  volume={35},
  pages={29776--29788},
  year={2022}
}

@inproceedings{zhang2019raven,
  title={RAVEN: A dataset for relational and analogical visual reasoning},
  author={Zhang, Chi and Gao, Feng and Jia, Baoxiong and Zhu, Yixin and Zhu, Song-Chun},
  booktitle={CVPR},
  pages={5317--5327},
  year={2019}
}

@inproceedings{barrett2018measuring,
  title={Measuring abstract reasoning in neural networks},
  author={Barrett, David and Hill, Felix and Santoro, Adam and Morcos, Ari and Lillicrap, Timothy},
  booktitle={ICML},
  pages={511--520},
  year={2018}
}

@article{fleuret2011comparing,
  title={Comparing machines and humans on a visual categorization test},
  author={Fleuret, Fran{\c{c}}ois and Li, Ting and Dubout, Charles and Wampler, Emma K and Yantis, Steven and Geman, Donald},
  journal={PNAS},
  volume={108},
  number={43},
  pages={17621--17625},
  year={2011},
  publisher={National Acad Sciences}
}

@article{wu2020scattering,
  title={The scattering compositional learner: Discovering objects, attributes, relationships in analogical reasoning},
  author={Wu, Yuhuai and Dong, Honghua and Grosse, Roger and Ba, Jimmy},
  journal={arXiv preprint, arXiv:2007.04212},
  year={2020}
}

@article{malkinski2022review,
title = {A review of emerging research directions in Abstract Visual Reasoning},
journal = {Information Fusion},
volume = {91},
pages = {713-736},
year = {2023},
issn = {1566-2535},
author = {Mikołaj Małkiński and Jacek Mańdziuk},
}

@inproceedings{he2023hierarchical,
  title={Hierarchical {ConViT} with attention-based relational reasoner for visual analogical reasoning},
  author={He, Wentao and Zhang, Jialu and Ren, Jianfeng and Bai, Ruibin and Jiang, Xudong},
  booktitle={AAAI},
  volume={37},
  number={1},
  pages={22--30},
  year={2023}
}

@article{ricci2021same,
  title={Same-different conceptualization: a machine vision perspective},
  author={Ricci, Matthew and Cad{\`e}ne, R{\'e}mi and Serre, Thomas},
  journal={Curr. Opin. Behav. Sci},
  volume={37},
  pages={47--55},
  year={2021},
  publisher={Elsevier}
}

@inproceedings{shanahan2020explicitly,
  title={An explicitly relational neural network architecture},
  author={Shanahan, Murray and Nikiforou, Kyriacos and Creswell, Antonia and Kaplanis, Christos and Barrett, David and Garnelo, Marta},
  booktitle={ICML},
  pages={8593--8603},
  year={2020}
}

@inproceedings{benny2021scale,
  title={Scale-localized abstract reasoning},
  author={Benny, Yaniv and Pekar, Niv and Wolf, Lior},
  booktitle={CVPR},
  pages={12557--12565},
  year={2021}
}

@inproceedings{yang2023neural,
  title={Neural prediction errors enable analogical visual reasoning in human standard intelligence tests},
  author={Yang, Lingxiao and You, Hongzhi and Zhen, Zonglei and Wang, Dahui and Wan, Xiaohong and Xie, Xiaohua and Zhang, Ru-Yuan},
  booktitle={ICML},
  pages={39572--39583},
  year={2023}
}

@inproceedings{chen2020simple,
  title={A simple framework for contrastive learning of visual representations},
  author={Chen, Ting and Kornblith, Simon and Norouzi, Mohammad and Hinton, Geoffrey},
  booktitle={ICML},
  pages={1597--1607},
  year={2020}
}

@inproceedings{chen2021and,
  title={An empirical study of training self-supervised vision transformer},
  author={Chen, X and Xie, S and He, K},
  booktitle={ICCV},
  pages={9620--9629},
  year={2021}
}

@article{oord2018representation,
  title={Representation learning with contrastive predictive coding},
  author={Oord, Aaron van den and Li, Yazhe and Vinyals, Oriol},
  journal={arXiv preprint arXiv:1807.03748},
  year={2018}
}

@inproceedings{robinson2020contrastive,
  title={Contrastive Learning with Hard Negative Samples},
  author={Robinson, Joshua David and Chuang, Ching-Yao and Sra, Suvrit and Jegelka, Stefanie},
  booktitle={ICLR},
  year={2020}
}

@inproceedings{chuang2022robust,
  title={Robust contrastive learning against noisy views},
  author={Chuang, Ching-Yao and Hjelm, R Devon and Wang, Xin and Vineet, Vibhav and Joshi, Neel and Torralba, Antonio and Jegelka, Stefanie and Song, Yale},
  booktitle={CVPR},
  pages={16670--16681},
  year={2022}
}

@inproceedings{jiang2023mixphm,
  title={MixPHM: Redundancy-Aware Parameter-Efficient Tuning for Low-Resource Visual Question Answering},
  author={Jiang, Jingjing and Zheng, Nanning},
  booktitle={CVPR},
  pages={24203--24213},
  year={2023}
}

@article{allen2020rapid,
  title={Rapid trial-and-error learning with simulation supports flexible tool use and physical reasoning},
  author={Allen, Kelsey R and Smith, Kevin A and Tenenbaum, Joshua B},
  journal={PNAS},
  volume={117},
  number={47},
  pages={29302--29310},
  year={2020},
  publisher={National Acad Sciences}
}

@inproceedings{baradel2019cophy,
  title={COPHY: Counterfactual Learning of Physical Dynamics},
  author={Baradel, Fabien and Neverova, Natalia and Mille, Julien and Mori, Greg and Wolf, Christian},
  booktitle={ICLR},
  year={2020}
}

@article{dai2019bridging,
  title={Bridging machine learning and logical reasoning by abductive learning},
  author={Dai, Wang-Zhou and Xu, Qiuling and Yu, Yang and Zhou, Zhi-Hua},
  journal={NeurIPS},
  volume={32},
  year={2019}
}

@inproceedings{hu2021stratified,
  title={Stratified rule-aware network for abstract visual reasoning},
  author={Hu, Sheng and Ma, Yuqing and Liu, Xianglong and Wei, Yanlu and Bai, Shihao},
  booktitle={AAAI},
  volume={35},
  number={2},
  pages={1567--1574},
  year={2021}
}

@article{zador2023catalyzing,
  title={Catalyzing next-generation artificial intelligence through {neuroAI}},
  author={Zador, Anthony and Escola, Sean and Richards, Blake and {\"O}lveczky, Bence and Bengio, Yoshua and Boahen, Kwabena and Botvinick, Matthew and Chklovskii, Dmitri and Churchland, Anne and Clopath, Claudia and others},
  journal={Nat. Commun.},
  volume={14},
  pages={1597},
  doi={10.1038/s41467-023-37180-x},
  year={2023}
}

@inproceedings{kirillov2023segment,
  title={Segment anything},
  author={Kirillov, Alexander and Mintun, Eric and Ravi, Nikhila and Mao, Hanzi and Rolland, Chloe and Gustafson, Laura and Xiao, Tete and Whitehead, Spencer and Berg, Alexander C and Lo, Wan-Yen and others},
  booktitle={CVPR},
  pages={4015--4026},
  year={2023}
}

@inproceedings{saxton2019analysing,
  title={Analysing Mathematical Reasoning Abilities of Neural Models},
  author={Saxton, David and Grefenstette, Edward and Hill, Felix and Kohli, Pushmeet},
  booktitle={ICLR},
  year={2018}
}

@inproceedings{lake2018generalization,
  title={Generalization without systematicity: On the compositional skills of sequence-to-sequence recurrent networks},
  author={Lake, Brenden and Baroni, Marco},
  booktitle={ICML},
  pages={2873--2882},
  year={2018}
}

@inproceedings{thrush2022winoground,
  title={Winoground: Probing vision and language models for visio-linguistic compositionality},
  author={Thrush, Tristan and Jiang, Ryan and Bartolo, Max and Singh, Amanpreet and Williams, Adina and Kiela, Douwe and Ross, Candace},
  booktitle={CVPR},
  pages={5238--5248},
  year={2022}
}

@inproceedings{son2022contrastive,
  title={Contrastive learning for space-time correspondence via self-cycle consistency},
  author={Son, Jeany},
  booktitle={CVPR},
  pages={14679--14688},
  year={2022}
}

@article{tian2020makes,
  title={What makes for good views for contrastive learning?},
  author={Tian, Yonglong and Sun, Chen and Poole, Ben and Krishnan, Dilip and Schmid, Cordelia and Isola, Phillip},
  journal={NeurIPS},
  volume={33},
  pages={6827--6839},
  year={2020}
}

@inproceedings{deng2009imagenet,
  title={ImageNet: A large-scale hierarchical image database},
  author={Deng, Jia and Dong, Wei and Socher, Richard and Li, Li-Jia and Li, Kai and Fei-Fei, Li},
  booktitle={CVPR},
  pages={248--255},
  year={2009}
}

@inproceedings{radford2021learning,
  title={Learning transferable visual models from natural language supervision},
  author={Radford, Alec and Kim, Jong Wook and Hallacy, Chris and Ramesh, Aditya and Goh, Gabriel and Agarwal, Sandhini and Sastry, Girish and Askell, Amanda and Mishkin, Pamela and Clark, Jack and others},
  booktitle={ICML},
  pages={8748--8763},
  year={2021}
}

@article{kasneci2023chatgpt,
  title={{ChatGPT} for good? On opportunities and challenges of large language models for education},
  author={Kasneci, Enkelejda and Se{\ss}ler, Kathrin and K{\"u}chemann, Stefan and Bannert, Maria and Dementieva, Daryna and Fischer, Frank and Gasser, Urs and Groh, Georg and G{\"u}nnemann, Stephan and H{\"u}llermeier, Eyke and others},
  journal={Learn. Individ. Differ.},
  volume={103},
  pages={102274},
  year={2023},
  publisher={Elsevier}
}

@inproceedings{camposampiero2023abstract,
  title={Abstract visual reasoning enabled by language},
  author={Camposampiero, Giacomo and Houmard, Lo{\"\i}c and Estermann, Benjamin and Mathys, Jo{\"e}l and Wattenhofer, Roger},
  booktitle={CVPR},
  pages={2642--2646},
  year={2023}
}

@inproceedings{li2024regression,
  title={Regression residual reasoning with pseudo-labeled contrastive learning for uncovering multiple complex compositional relations},
  author={Li, Chengtai and He, Yuting and Ren, Jianfeng and Bai, Ruibin and Zhao, Yitian and Yu, Heng and Jiang, Xudong},
  booktitle={IJCAI},
  volume={4},
  pages={3466--3474},
  year={2024}
}

@inproceedings{li2025darr,
  title={{DARR}: A dual-branch arithmetic regression reasoning framework for solving machine number reasoning},
  author={Li, Chengtai and Tan, Yee Yang and He, Yuting and Ren, Jianfeng and Bai, Ruibin and Zhao, Yitian and Yu, Heng and Jiang, Xudong},
  booktitle={AAAI},
  volume={39},
  number={2},
  pages={1373--1382},
  year={2025}
}

@inproceedings{malkinski2024one,
  title={One self-configurable model to solve many abstract visual reasoning problems},
  author={Ma{\l}ki{\'n}ski, Miko{\l}aj and Ma{\'n}dziuk, Jacek},
  booktitle={AAAI},
  volume={38},
  number={13},
  pages={14297--14305},
  year={2024}
}

@ARTICLE{song2024centerformer,
  author={Song, Wenfeng and Wang, Xuan and Guo, Yuting and Li, Shuai and Xia, Bin and Hao, Aimin},
  journal={TMM}, 
  title={{CenterFormer}: A Novel Cluster Center Enhanced Transformer for Unconstrained Dental Plaque Segmentation}, 
  year={2024},
  volume={26},
  number={},
  pages={10965-10978},
  doi={10.1109/TMM.2024.3428349}}

@ARTICLE{umam2024unsupervised,
  author={Umam, Ardian and Yang, Cheng-Kun and Chuang, Jen-Hui and Lin, Yen-Yu},
  journal={TMM}, 
  title={Unsupervised Point Cloud Co-Part Segmentation via Co-Attended Superpoint Generation and Aggregation}, 
  year={2024},
  volume={26},
  number={},
  pages={7775-7786},
  doi={10.1109/TMM.2024.3371294}}

@ARTICLE{zhu2024multi,
  author={Zhu, Jian and Wang, Hanli and He, Bin},
  journal={TMM}, 
  title={Multi-Modal Structure-Embedding Graph Transformer for Visual Commonsense Reasoning}, 
  year={2024},
  volume={26},
  number={},
  pages={1295-1305},
  doi={10.1109/TMM.2023.3279691}}

@ARTICLE{pan2024joint,
  author={Pan, Renjie and Yang, Hua and Li, Cunyan and Yang, Jinhai},
  journal={TMM}, 
  title={Joint Intra \& Inter-Grained Reasoning: A New Look Into Semantic Consistency of Image-Text Retrieval}, 
  year={2024},
  volume={26},
  number={},
  pages={4912-4925},
  doi={10.1109/TMM.2023.3327645}}

@inproceedings{li2025dbcr,
  title={{DBCR}: Exploiting Both Intra-cluster and Extra-cluster Relations for Compositional Reasoning},
  author={Li, Chengtai and Su, Guosheng and Ren, Jianfeng and Bai, Ruibin and Zhao, Yitian and Jiang, Xudong},
  booktitle={ICASSP},
  pages={1--5},
  year={2025}
}

@inproceedings{MRNet,
  title={Scale-localized abstract reasoning},
  author={Benny, Yaniv and Pekar, Niv and Wolf, Lior},
  booktitle={CVPR},
  pages={12557--12565},
  year={2021}
}

@article{lake2017building,
title={Building machines that learn and think like people}, 
volume={40}, 
DOI={10.1017/S0140525X16001837}, 
journal={Behav. Brain Sci.}, 
author={Lake, Brenden M. and Ullman, Tomer D. and Tenenbaum, Joshua B. and Gershman, Samuel J.}, 
year={2017}, 
pages={e253}
}

@ARTICLE{małkiński2024multi,
  author={Małkiński, Mikołaj and Mańdziuk, Jacek},
  journal={TNNLS}, 
  title={Multi-Label Contrastive Learning for Abstract Visual Reasoning}, 
  year={2024},
  volume={35},
  number={2},
  pages={1941-1953},
  doi={10.1109/TNNLS.2022.3185949}}

@book{nie2022learning,
  title={Learning from multiple social networks},
  author={Nie, Liqiang and Song, Xuemeng and Chua, Tat-Seng},
  year={2022},
  publisher={Springer Nature}
}

@inproceedings{nie2022search,
  title={Search-oriented micro-video captioning},
  author={Nie, Liqiang and Qu, Leigang and Meng, Dai and Zhang, Min and Tian, Qi and Bimbo, Alberto Del},
  booktitle={ACM MM},
  pages={3234--3243},
  year={2022}
}

@article{he2025two,
title = {Two-stage Rule-induction visual reasoning on {RPMs} with an application to video prediction},
journal = PR,
volume = {160},
pages = {111151},
year = {2025},
issn = {0031-3203},
author = {Wentao He and Jianfeng Ren and Ruibin Bai and Xudong Jiang},
}

@article{he2024data,
title = {Data augmentation by morphological mixup for solving {Raven's} progressive matrices},
journal = {Vis Comput},
volume = {40},
pages = {2457-2470},
year = {2024},
author = {Wentao He and Jianfeng Ren and Ruibin Bai},
}

@inproceedings{he2024hierarchical,
  title={Hierarchical Perceptual and Predictive Analogy-Inference Network for Abstract Visual Reasoning},
  author={Wentao He and Jianfeng Ren and Ruibin Bai and Xudong Jiang},
  booktitle={ACM MM},
  pages={4841-4850},
  year={2024}
}

@INPROCEEDINGS{wu2025mg,
  author={Wu, Bizhu and Xie, Jinheng and Shen, Keming and Kong, Zhe and Ren, Jianfeng and Bai, Ruibin and Qu, Rong and Shen, Linlin},
  booktitle= {CVPR}, 
  title={{MG-MotionLLM}: A Unified Framework for Motion Comprehension and Generation across Multiple Granularities}, 
  year={2025},
  pages={27849-27858},
}

@inproceedings{zhang2024visual,
author = {Zhang, Jialu and Wang, Xinyi and Yao, Chenglin and Ren, Jianfeng and Jiang, Xudong},
title = {Visual-linguistic Cross-domain Feature Learning with Group Attention and Gamma-correct Gated Fusion for Extracting Commonsense Knowledge},
year = {2024},
booktitle = {ACM MM},
pages = {4650–4659},
numpages = {10},
}

@ARTICLE{xue2025linin,
  author={Xue, Dizhan and Qian, Shengsheng and Fang, Quan and Xu, Changsheng},
  journal= {TMM},
  title={{LININ}: Logic Integrated Neural Inference Network for Explanatory Visual Question Answering}, 
  year={2025},
  volume={27},
  number={},
  pages={16-27},
}

@ARTICLE{yuan2025relation,
  author={Yuan, Mengqi and Jia, Gengyun and Bao, Bing-Kun},
  journal={TMM}, 
  title={Relation Inference Enhancement Network for Visual Commonsense Reasoning}, 
  year={2025},
  volume={27},
  number={},
  pages={2221-2231},
}

@ARTICLE{xiong20253urllm,
  author={Xiong, Haomiao and Zhuge, Yunzhi and Zhu, Jiawen and Zhang, Lu and Lu, Huchuan},
  journal={TMM}, 
  title={{3UR-LLM}: An End-to-End Multimodal Large Language Model for {3D} Scene Understanding}, 
  year={2025},
  volume={27},
  number={},
  pages={2899-2911},
}

@inproceedings{he2024dual,
  title={Dual-Branch StarNet with Mutual Attention and U-Net Denoising for Simultaneously Recognizing Keywords and Speakers},
  author={He, Yuting and Li, Chengtai and Yu, Heng and Ren, Jianfeng and Wang, Zheng and Du, Heshan and Xia, Yinshui},
  booktitle={ICONIP},
  pages={286--300},
  year={2024},
  organization={Springer}
}

@inproceedings{lidsrf,
  title={DSRF: A Dynamic and Scalable Reasoning Framework for Solving RPMs},
  author={Li, Chengtai and He, Yuting and Ren, Jianfeng and Bai, Ruibin and Zhao, Yitian and Jiang, Xudong},
  booktitle={NeurIPS},
  year={2025}
}

\end{document}